\documentclass{article}

 \usepackage[main, final]{neurips_2026}
\usepackage[utf8]{inputenc} % allow utf-8 input
\usepackage[T1]{fontenc}    % use 8-bit T1 fonts
\usepackage{hyperref}       % hyperlinks
\usepackage{url}            % simple URL typesetting
\usepackage{booktabs}       % professional-quality tables
\usepackage{amsfonts}       % blackboard math symbols
\usepackage{nicefrac}       % compact symbols for 1/2, etc.
\usepackage{microtype}      % microtypography
\usepackage{xcolor}         % colors

\usepackage{graphicx}
\usepackage{booktabs}
\usepackage{tabularx}
\definecolor{darkgreen}{RGB}{0, 100, 0}

\newcommand{\sgclsp}{\ensuremath{\text{SGCls}^+}\xspace}
\usepackage{multirow}
\usepackage{tabularx}
\usepackage{array}
\usepackage{colortbl}
\usepackage{xcolor}
\usepackage{soul}
\usepackage{makecell}
\usepackage{url}
\usepackage{seqsplit}
\usepackage{graphicx}
\usepackage{booktabs}
\usepackage{pifont}
\usepackage{svg}
\usepackage{amssymb}
\usepackage{amsmath}
\usepackage{caption}
\usepackage{graphicx}
\usepackage{pifont}
\usepackage{enumitem}
\usepackage{float}
\usepackage[table]{xcolor}
\usepackage{subcaption}
\usepackage{xspace}
\usepackage{wrapfig}
\newcommand{\cmark}{\ding{51}} % ✓
\newcommand{\xmark}{\ding{55}}

\definecolor{lightergray}{rgb}{0.97, 0.97, 0.97}
\definecolor{lblue}{HTML}{F3F2FC}
\definecolor{lgreen}{HTML}{EDF7ED}

\newcommand{\inc}[1]{%
{\scriptsize\textcolor{green!40!black}{(+#1)}}%
}

\newcommand{\dec}[1]{%
{\scriptsize\textcolor{red!30!black}{(-#1)}}%
}

\usepackage[accsupp]{axessibility}  % Improves PDF readability for those with disabilities.

\usepackage{hyperref}

\usepackage{orcidlink}
\hypersetup{
    colorlinks=true,
    citecolor=black,
    linkcolor=black,
    urlcolor=black
}

\begin{document}

\title{A Closer Look at Dynamic Scene Graph Generation \\ In the Era of \\Multimodal Large Language Models} 

% \author{%
%   David S.~Hippocampus\thanks{Use footnote for providing further information
%     about author (webpage, alternative address)---\emph{not} for acknowledging
%     funding agencies.} \\
%   Department of Computer Science\\
%   Cranberry-Lemon University\\
%   Pittsburgh, PA 15213 \\
%   \texttt{hippo@cs.cranberry-lemon.edu} \\
%   % examples of more authors
%   % \And
%   % Coauthor \\
%   % Affiliation \\
%   % Address \\
%   % \texttt{email} \\
%   % \AND
%   % Coauthor \\
%   % Affiliation \\
%   % Address \\
%   % \texttt{email} \\
%   % \And
%   % Coauthor \\
%   % Affiliation \\
%   % Address \\
%   % \texttt{email} \\
%   % \And
%   % Coauthor \\
%   % Affiliation \\
%   % Address \\
%   % \texttt{email} \\
% }

\author{%
  {\small Xuanming Cui\textsuperscript{1} \enspace
  Jaiminkumar Ashokbhai Bhoi\textsuperscript{1} \enspace
  Chionh Wei Peng\textsuperscript{2} \enspace
  Adriel Kuek\textsuperscript{2} \enspace
  Ser Nam Lim\textsuperscript{1}} \\[1.5ex]
  \normalfont \textsuperscript{1}University of Central Florida \quad
  \textsuperscript{2}DSO National Laboratories \\
  \texttt{\{xuanming.cui, ja882177, sernam\}@ucf.edu} \\
  \texttt{\{weipeng, kyongjie\}@dso.org.sg}
}

\maketitle

\begin{abstract}

Dynamic Scene Graph Generation (DSGG) aims to capture objects and their evolving relations in videos. Despite recent progress, the practicality and quality of generated scene graphs remain limited compared to the rapid advances in Multimodal Large Language Models (MLLMs). In this work, we revisit DSGG from two fundamental perspectives:  task setup and model design. From the task setup perspective, we identify two key limitations of the current recall-oriented evaluation protocol: (i) a severe precision–recall trade-off, and (ii) uninformative and redundant relation generation. To better assess practical usefulness, we introduce five additional metrics that provide a more comprehensive evaluation of the quality of generated dynamic scene graphs. From the model design perspective, we explore directly using MLLMs for scene graph generation and establish a strong MLLM-based DSGG baseline through three design changes. First, we replace the conventional bottom-up pipeline with a top-down \emph{reason-then-locate} strategy. Second, we reformulate frame-wise dynamic graphs as {Temporal Relation Set (TRS)} prediction, improving both efficiency and performance. Third, we introduce {Importance-Aware Finetuning (IAF)} to encourage more relevant and diverse relation generation. Extensive experiments on Action Genome, VidVRD, and PVSG show that our approach consistently achieves state-of-the-art performance.

% The ABSTRACT is to be in fully justified italicized text, at the top of the left-hand column, below the author and affiliation information.
% Use the word ``Abstract'' as the title, in 12-point Times, boldface type, centered relative to the column, initially capitalized.
% The abstract is to be in 10-point, single-spaced type.
% Leave two blank lines after the Abstract, then begin the main text.
% Look at previous \confName abstracts to get a feel for style and length.
\end{abstract}    
\section{Introduction}
\label{sec:intro}

Understanding complex visual scenes requires not only recognizing objects but also reasoning about the relationships between them. Scene Graph Generation (SGG) addresses this problem by representing visual content as a structured graph where nodes correspond to objects and edges describe their relations. Such structured representations provide an interpretable intermediate representation and have been shown to benefit a range of downstream tasks, including image retrieval~\cite{cherian2022251dspatiotemporalscenegraphs,10.1109/TMM.2022.3169065,mao-etal-2022-dynamic,lei2022symbolicreplayscenegraph}, visual question answering (VQA)~\cite{cherian2022251dspatiotemporalscenegraphs,10.1109/TMM.2022.3169065,mao-etal-2022-dynamic,lei2022symbolicreplayscenegraph}, and controllable image generation~\cite{farshad2023scenegeniescenegraphguided}. On top of static image-based SGG, Dynamic Scene Graph Generation (DSGG)~\cite{sttran, tempura, oed} aims to capture both objects and their evolving relationships over time, enabling structured understanding of dynamic scenes through temporal dimension.

Despite steady progress in DSGG research, the practical quality of generated scene graphs remains limited, compared to the wide adaptation of Multimodal Large Language Models (MLLMs). Recent advances in MLLMs have dramatically improved visual understanding capabilities, demonstrating strong performance on a wide range of practical visual tasks such as video captioning, reasoning~\cite{qwen3vl, gemini2_5}, and retrieval~\cite{tte}. This contrast raises a natural question: 

\vspace{-5pt}
\begin{center}\emph{Why does DSGG lag behind the broader progress in multimodal video understanding?}
\end{center}
\vspace{-10pt}

In this work, we re-examine the task of DSGG to better understand this discrepancy from two fundamental perspectives: \emph{1) task setup}, and \emph{2) model design}.

\textbf{Rethinking DSGG Task Setup.} Existing DSGG methods are evaluated exclusively using recall-based metrics. We argue that recall alone provides an incomplete view of scene graph quality.

\begin{figure}
\centering
\begin{minipage}[b]{0.47\linewidth}
\centering
\includegraphics[width=\linewidth]{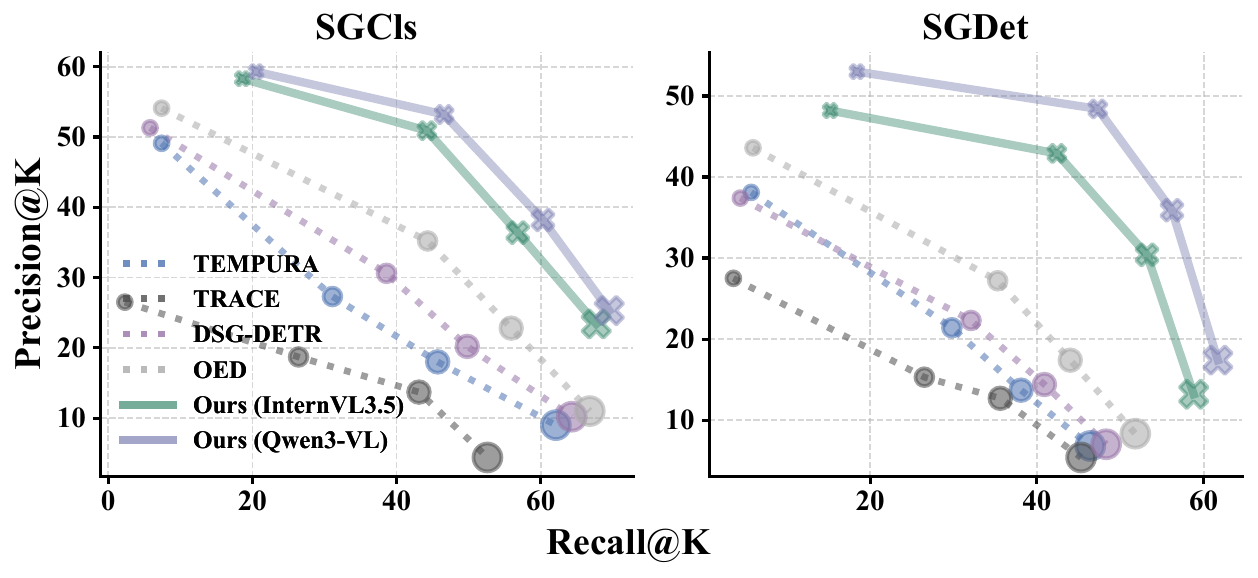}
\vspace{-10pt}
\caption{ Precision-Recall~\protect\footnotemark curve for previous DSGG~\cite{sttran, tempura, oed} and our MLLM-based methods for two DSGG tasks (SGCls and SGDet) on Action Genome~\cite{ji2020actiongenome}. We plot top-$k$ with $k \in \{1,10,20,50\}$, indicated by the size of the markers. We observe severe precision-recall trade-off on existing methods such as OED~\cite{oed}, as opposed to our MLLM-based methods, whose precision are notably better.}
\label{fig:pre-recall}
\end{minipage}
\vspace{-15pt}
\hfill
\begin{minipage}[b]{0.52\linewidth}
\centering
\includegraphics[width=\linewidth]{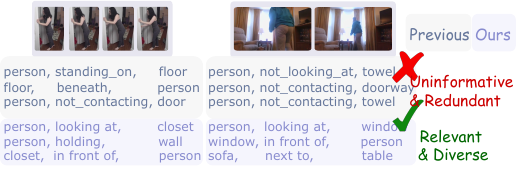}
\captionof{figure}{ Visualization of top-3 prediction comparison with previous DSGG method~\cite{oed} and our MLLM-based method. Two sample video clips are from Action Genome~\cite{ji2020actiongenome}. While previous DSGG method predicts correct but uninformative and redundant triplets  (e.g. $\langle \texttt{person}, \texttt{standing\_on}, \texttt{floor} \rangle$ and $\langle \texttt{floor}, \texttt{beneath}, \texttt{person} \rangle$), ours predicts $\langle \texttt{person}, \texttt{looking\_at}, \texttt{closet} \rangle$, which is the primary activity of the scene, while covering more semantics of the scene.}
\label{fig:uninformative}

\end{minipage}
\vspace{-15pt}
\end{figure}

\textit{Precision-Recall Trade-off.} Recall-only evaluation ignores the precision–recall trade-off. Existing DSGG models can artificially inflate recall by predicting a large number of relations at the expense of prediction precision. This behavior frequently results in scene graphs that contain numerous incorrect and noisy relations. While such trade-off is prevalent in regular retrieval systems, it brings substantially more impact to DSGG, as the goal of DSGG is not retrieval, but to provide accurate spatial-temporal information for downstream applications such as video QA and robotic planning~\citep{esca}. In such scenarios, severely damaging the performance of downstream tasks (Figure~\ref{fig:pre-recall} and~\ref{fig:sgret_k}).

\textit{uninformative and redundant relation generation.} Existing literature typically views DSGG as an unordered retrieval problem, where the relative importance of predicted relations and the diversity of the generated scene graphs are ignored. However, not all relations contribute equally to understanding a scene. Semantically meaningful interactions (e.g., $\langle \texttt{person}, \texttt{riding}, \texttt{horse} \rangle$) are often far more informative than trivial spatial relations (e.g., $\langle \texttt{floor}, \texttt{beneath}, \texttt{person} \rangle$). Moreover, as uninformative relations (e.g. \texttt{not looking at}) are more prevalent in training data~\cite{ji2020actiongenome}, traditional DSGG methods tend to produce higher probability for such relations, resulting in uninformative relations dominating top-ranked predictions (Figure~\ref{fig:uninformative}). While this may seem relevant to the well-studied long-tail problem in DSGG, we note that whether a relation is uninformative or redundant depends on the video and generated relations. A long-tail relation (\textit{e.g.} \texttt{``have it on the back''}) can be uninformative, whereas a common relation (\textit{e.g.} \texttt{``in''}) may still be informative. 

\footnotetext{While MLLM-based DSGG's precision also seems to degrade as recall increases, this is due to how evaluation is performed: we fill MLLM's predictions with random triplets when the number of output is less than $k$. In reality, due to the autoregressive nature, MLLMs \textit{``knows''} to stop generation when it does not find further plausible predictions.}

To better assess the practical utility of generated scene graphs, we introduce five complementary metrics. In addition to precision, we introduce \textbf{Importance Aware Ranking}, which treat DSGG as a ranking task to consider triplet importance and diversity. We further incorporate two downstream tasks as evaluation metrics: \textbf{Scene Graph Retrieval} (SGRet), which measures the effectiveness of scene graphs via video retrieval, and \textbf{Scene Graph Question Answering} (SGQA), a curated Multi Choice Question (MCQ) dataset designed to evaluate reasoning over generated scene graphs. Lastly, we propose \textbf{\sgclsp}, a modified DSGG task setting, which provides more realistic assessment of relation predictions. Together, these metrics provide a more comprehensive assessment of DSGG quality beyond the widely-used recall-based metrics~\footnote{Note that while the above discussion could also apply to image-based SGG, we focus on the video modality as it is substantially more complex than image-based SGG, and has wider downstream applications.}.

\textbf{Rethinking DSGG Model Design.} Most prior DSGG approaches follow a bottom-up pipeline inherited from image-based SGG: objects are first detected, the regional features are subsequently passed through specialized spatial–temporal aggregation modules to infer object and relation types. While effective to some extent, this design introduces several challenges.

\textbf{(1)} Bottom-up pipelines are inherently bottlenecked by the object detector, as relation prediction relies entirely on detected regions. Errors or missing detections can propagate through the pipeline and degrade final scene graph quality. \textbf{(2)} Existing DSGG models~\cite{sttran, trace, dsg_detr, oed} typically rely on hand-crafted spatial–temporal modules, leading to increasingly complex architectures that are difficult to generalize across datasets or tasks. \textbf{(3)} Detector-based pipelines typically operate on cropped object regions, which may discard important contextual information needed to understand interactions in a scene. \textbf{(4)} Existing DSGG models are mostly trained from scratch for each individual task, limiting their ability to leverage broader knowledge from large-scale multimodal data. 

In contrast, modern MLLMs adopt a fundamentally different paradigm for visual reasoning. Rather than building visual understanding from localized detections, MLLMs perform holistic reasoning over the entire visual input, leveraging large-scale pretraining and multimodal knowledge. This top-down process better resembles how humans interpret visual scenes, where interactions and activities are often understood before individual components are explicitly localized. Such capabilities suggest that MLLMs may provide a promising alternative foundation for scene graph generation.

% \noindent\textit{Towards MLLM-based Dynamic Scene Graph Generation.} Despite the remarkable progress of MLLMs in video understanding, their capability for dense relational prediction has not been systematically studied. Existing work involving MLLMs in DSGG research has primarily focused on generating synthetic training data~\cite{laser, kim2025weaklysupervisedvideoscene, esca}, rather than directly using MLLMs to generate scene graphs themselves. Consequently, it remains unclear to what extent can MLLMs effectively support dense, structured video prediction tasks such as DSGG.

Motivated by the above observation, we provide the first comprehensive study of adopting MLLMs for direct video scene graph generation. We first reveal that while SOTA MLLMs excel at single prediction verification (Table~\ref{tab:zs_verification_acc}), they do not naturally work for dense visual generation (Figure~\ref{fig:zs_recall_comparison}). Subsequently, we introduce three novel design changes that adapt MLLMs to the DSGG task. \textbf{(1)} We depart from the traditional bottom-up pipeline and adopt a top-down \emph{reason-then-locate} strategy, where interactions are first inferred semantically before being grounded in objects. \textbf{(2)} We reformulate frame-wise DSGG as Temporal Relation Set (TRS) prediction, which improves both efficiency and temporal coherence for MLLM-based DSGG. \textbf{(3)} We propose Importance-Aware Finetuning (IAF) to guide the model toward generating ranked relations that are both relevant and diverse.

% We provide a comprehensive evaluation of the MLLM-based DSGG approach under various datasets and tasks. Our experiments demonstrate that our proposed MLLM-based baseline consistently outperforms existing DSGG methods across both traditional and newly proposed evaluation metrics by a substantial margin, highlighting the limitations of recall-oriented task setups and the improved practical utility of the scene graphs generated by our method.

Our contributions are summarized as follows:

\noindent\textbf{{1) Rethinking DSGG Task Setup.}} We identify key limitations of the current recall-based evaluation protocol and introduce five complementary metrics that better reflect the practical quality of generated scene graphs. In addition, we curate a high-quality MCQ dataset to effectively evaluate SGQA.

\noindent\textbf{{2) Rethinking DSGG model design.}} We are the first to formalize a \textit{Reason-then-Locate} paradigm for DSGG. To effectively adapt MLLM as scene graph generators, we propose TRS and IAF to improve the performance, efficiency, and quality of the predicted scene graphs.

\noindent\textbf{{3) A strong baseline for MLLM-based DSGG.}} We provide the first systematic study of adopting MLLMs for direct DSGG. Through extensive evaluation across multiple datasets, tasks, and metrics, our MLLM-based approach demonstrates significant improvements over existing SOTAs.

\section{Related Work}
\label{sec:related_works}

\paragraph{Scene Graph Generation.}

Scene Graph Generation (SGG) provides a graph-based representation of the visual content. Traditional approaches typically exploit the graph property of the scene graph via Message Passing~\cite{xu-2017, li2017vipcnn, yang2018graph, tang2018neural} and Graph Convolutional Network~\cite{gpsnet}. Another line of works focuses on handling the long-tailed problem due to coarse labeling~\cite{tang2020unbiasedscenegraphgeneration, sun2023unbiasedscenegraphgeneration, hilo}.
Given to the advances in Vision Language Models~\cite{clip, blip} and Multimodal Large Language Models~\cite{llava, qwen3vl}, several recent works utilize pre-trained multimodal foundational models for open-vocabulary SGG~\cite{vs3, pixel2graph}. For instance, $\text{VS}^3$ utilizes the pre-trained GLIP~\cite{glip} for region-word alignment, and \cite{pixel2graph} treats SGG as an image-to-sequence task by finetuning BLIP~\cite{blip} for entity/relation prediction and localization.

% \paragraph{GPT-Assisted SGG.} Another line of recent works utilizes Large Language Models (LLMs) to assist SGG. LLM4SGG~\cite{llm4sgg} utilizes ChatGPT to generate scene graph data from image (video) captions to train existing SGG methods for weakly-supervised SGG. Similarly, LASER~\cite{laser} also leverages ChatGPT to form ground-truth labels, but it converts natural language to formal video specifications and guides the training via a neuro-symbolic alignment score. VLPrompt~\cite{vlprompt} prompts a LLM to generate likely predicates given subject-object pairs, which are used as auxiliary context for relation prediction.

% Instead of using LLMs as an auxiliary module, in this work, we explore the possibility of leveraging LMMs for direct scene graph generation.

\paragraph{Dynamic Scene Graph Generation.}

Dynamic Scene Graph Generation (DSGG) extends SGG with the additional temporal dimension. Existing methods in DSGG typically resemble similar bottom-up architectural designs in traditional SGG models. For instance, STTran~\cite{sttran} adds an attention module as the temporal decoder on top of the spatial encoder. OED~\cite{oed} follows the similar architectural design but replace Faster-RCNN with DETR~\cite{detr} for one-stage DSGG. A few other works~\cite{tempura, xu2022metaspatiotemporaldebiasingvideo, khandelwal2024flocodeunbiaseddynamicscene} focus on the long-tail problem in DSGG. On the other hand, several recent works~\cite{laser, nag2025conformalpredictionmllmaided, nguyen2025hyperglmhypergraphvideoscene, esca} explore utilizing MLLMs to enhance DSGG performance. These works typically exploit MLLM as synthetic data generator for weakly-supervised DSGG~\cite{laser, esca}, as a post-processing tool~\cite{nag2025conformalpredictionmllmaided}, or as a downstream application module~\cite{nguyen2025hyperglmhypergraphvideoscene}. For instance, HyperGLM~\citep{nguyen2025hyperglmhypergraphvideoscene} generates video scene graph with traditional bottom-up approach, and pass the generated scene graph to text-only LLM for reasoning; LASER~\citep{laser} and ESCA~\citep{esca} utilizes MLLM for synthetic data generation. However, none of these previous works explore using MLLM directly as video scene graph generator. 

\paragraph{Multimodal Large Language Models.}
Following LLaVA~\cite{llava}'s success, a large amount of visual instruction-tuned Large Multimodal Models (MLLMs) have been developed~\cite{instructblip, minigpt4, llava1.5, llavanext}. These models are typically trained for captioning, standard VQA, object grounding, and  complex visual reasoning~\cite{qwen3vl, internvl3.5}. Despite the wide range of evaluation suites~\cite{mmmlu, video_mme, SOK-Bench}, there is a lack of assessment of MLLMs' capacity on fine-grained dense generation for video tasks such as DSGG.

\section{Method} \label{sec:methods}

\subsection{Problem Formulation}

Dynamic Scene Graph Generation (DSGG) aims to represent objects in a video and their interactions over time as a sequence of structured graphs. Given a video $V = \{I_t\}_{t=1}^{T}$ consisting of $T$ frames, the goal of DSGG is to predict a \emph{dynamic scene graph} $G = \{G_t\}_{t=1}^{T}$, where each frame-level graph contains a set of scene graph triplet $\langle \texttt{sub}, \texttt{rel}, \texttt{obj} \rangle$, where \texttt{sub}, \texttt{rel} and \texttt{obj} refer to subject, relation and object respectively. Compared to image scene graph generation, DSGG additionally requires modeling the temporal dynamics of object interactions.

\subsection{Evaluating the Practicability of DSGG}

Existing DSGG works primarily evaluate models using recall or mean recall. While useful for measuring coverage in retrieval tasks, we argue that recall alone does not provide a reliable assessment of the practical quality of generated scene graphs. We empirically observe two critical limitations: 

\textbf{(1) Precision--recall trade-off.} Existing DSGG methods often exhibit severe precision degradation as the number of sampled scene graph triplets increases (Figure~\ref{fig:pre-recall}). As a result, the generated graphs frequently contain a large number of inaccurate or noisy relations. Such noise significantly degrades the usefulness of scene graphs in downstream applications such as Video QA or retrieval (Figure~\ref{fig:sgret_k}).

\textbf{(2) Uninformative and redundant relations.} Due to the prevalence of trivial relations in existing DSGG datasets~\cite{ji2020actiongenome}, existing DSGG models tend to assign high confidence to trivial or repetitive relations (as each triplet prediction is made separately) (Figure~\ref{fig:uninformative}), leading to scene graphs that are unnecessarily large yet semantically weak. Such redundancy becomes problematic when scene graphs are used as compact video representations (e.g., for metadata storage) or when downstream algorithms are sensitive to graph size (e.g., graph neural networks or neuro-symbolic methods~\cite{scallop}).

To systematically investigate these issues, we introduce five additional metrics that evaluate DSGG from the perspective of practical utility.

\noindent\textbf{{(1) Precision with MLLM-as-a-Judge.}}
Ground-truth annotations often do not exhaustively cover all valid scene graph relations, making direct precision estimation difficult. To address this issue, we adopt an MLLM-as-a-Judge strategy to assess the validity of predicted relations. Specifically, we first remove predicted triplets that already appear in the ground-truth annotations. For the remaining predictions, we prompt a more powerful MLLM to determine whether each triplet represents a valid subject-object interaction within the corresponding temporal interval.

\begin{figure*}
    \centering

    % Left: Barplot
    \begin{minipage}[t]{0.52\textwidth}
        \centering
        \vspace{0pt}
        \includegraphics[width=\textwidth]{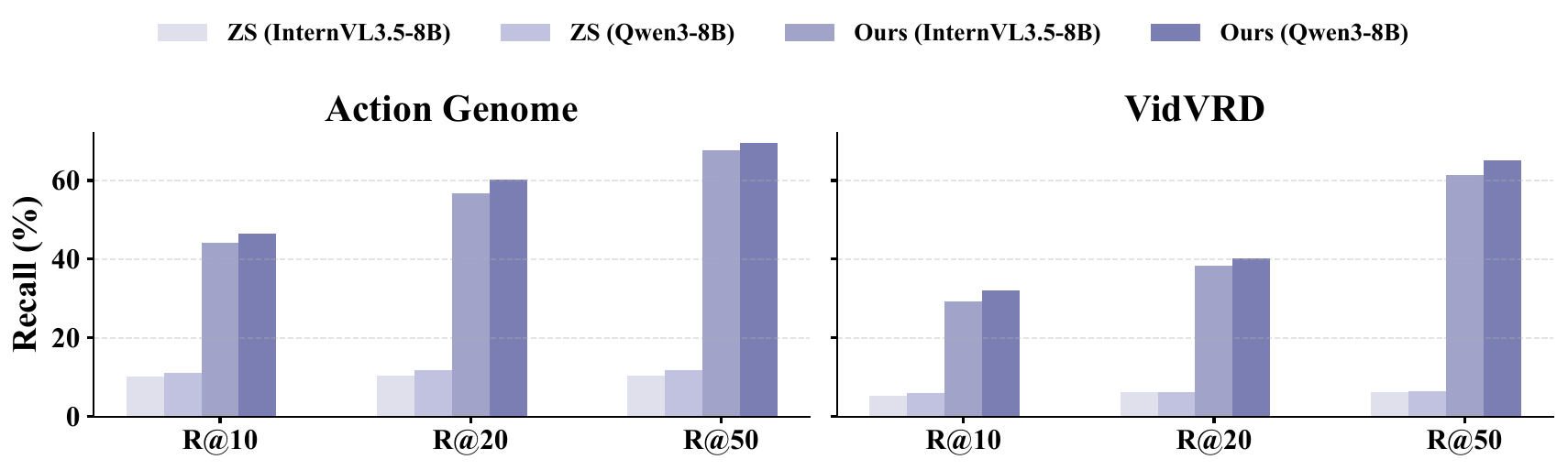}
        \vspace{-15pt}
        \caption{Recall comparison on \sgclsp. \textbf{ZS}: Zero-Shot prediction; \textbf{Ours}: MLLMs finetuned with our proposed procedure.}
        \label{fig:zs_recall_comparison}
    \end{minipage}
    \vspace{-5pt}
    \hfill % This pushes them to the far left and right
    % Note: No empty line here!
    \begin{minipage}[t]{0.45\textwidth}
        \centering
        \vspace{0pt}
        \captionof{table}{Individual scene graph triplet zero-shot verification performance. ``Truthful'' indicates accuracy on ground-truth triplets (``\texttt{yes}'' responses), and ``Adv.'' denotes rejection rate on adversarial triplets (``\texttt{no}'' responses). Higher is better ($\text{max} = 100$).}
        \vspace{-2pt}
        \resizebox{\linewidth}{!}{
            \begin{tabular}{lcc|cc}
            \toprule
            & \multicolumn{2}{c|}{ActionGenome}
            & \multicolumn{2}{c}{VidVRD} \\
            \cmidrule(lr){2-3} \cmidrule(lr){4-5}
            Method & Truthful (\%) & Adv. (\%) & Truthful (\%) & Adv. (\%) \\
            \midrule
            Zero-Shot (InternVL3.5-8B) & 96.4 & 98.2 & 98.6 & 98.1 \\
            Zero-Shot (Qwen3-8B)       & 97.0 & 98.5 & 98.3 & 99.5 \\
            \bottomrule
            \end{tabular}
        \label{tab:zs_verification_acc}
        }
    \end{minipage}
    \label{fig:main_results}
    \vspace{-5pt}
\end{figure*}

To validate the reliability of \emph{MLLM-as-a-Judge} over MLLM's generation, we benchmark MLLM's performance on verifying whether a given triplet is valid according to the video over 1K randomly sampled triplets from both Action Genome~\citep{ji2020actiongenome} and VidVRD~\citep{vidvrd}. We find that although MLLMs' zero-shot performance on dense DSGG is substantially lower than finetuned ones (Figure~\ref{fig:zs_recall_comparison}), their performance on individual triplet verification is exceptionally strong (Table~\ref{tab:zs_verification_acc}).  This confirms the reliability of utilizing \emph{MLLM-as-a-Judge} for DSGG. We further show that \sgclsp $P@10$ on AG is consistent across judges from different model families (Section~\ref{sec:analysis}).

\noindent{\textbf{(2) Scene Graph Retrieval (SGRet).}} We introduce \emph{Scene Graph Retrieval (SGRet)} to measure how effectively predicted graphs support video retrieval. Specifically, we convert predicted scene graphs into natural language descriptions and use an off-the-shelf multimodal embedding model (e.g., Qwen3VL-Embedding~\cite{qwen3vl_embedding}) to perform text-to-video retrieval. Retrieval accuracy then serves as a practical proxy for the semantic quality of generated scene graphs.

\begin{figure*}[t]
    \centering
    \includegraphics[width=\textwidth]{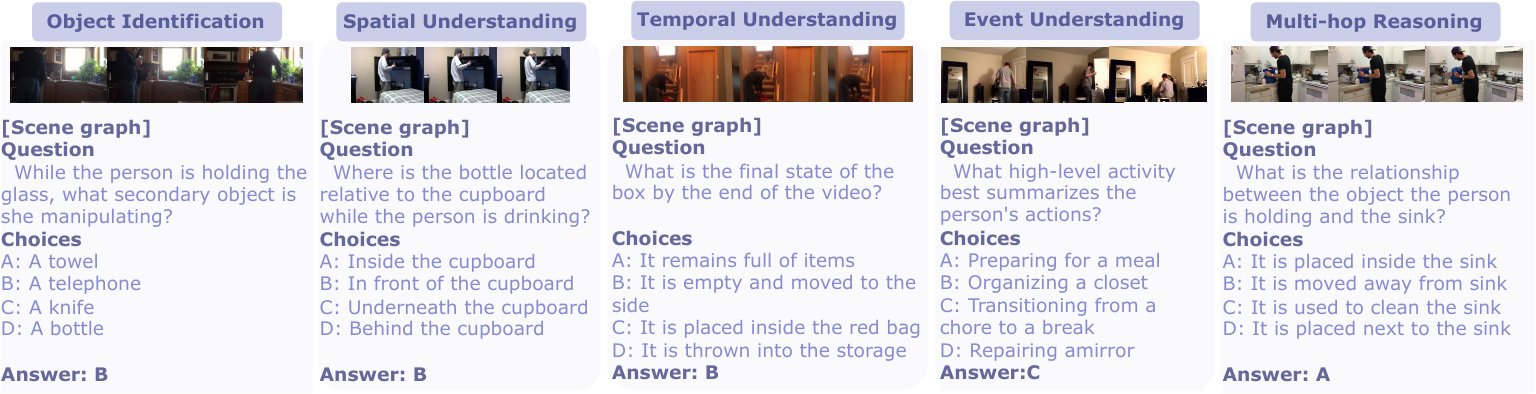}
    \caption{\footnotesize Samples of our curated SGQA dataset based on Action Genome.}
    \label{fig:sgqa_dataset}
\end{figure*}

\noindent{\textbf{(3) Scene Graph QA (SGQA).}} To assess the utility of generated scene graphs for downstream tasks, we introduce SGQA, a diagnostic benchmark comprising 1,000 multiple-choice questions derived from 200 Action Genome videos~\cite{ji2020actiongenome}. Unlike existing benchmarks like AGQA~\cite{agqa} that focus on compositional logic over fixed symbolic structures, SGQA employs a frontier MLLM (\textit{e.g.}, Gemini 2.5 Pro~\citep{gemini2_5}) to curate ``hard'' questions conditioned on both raw video and ground-truth scene graphs.

We focus on five reasoning dimensions: object/action identification, spatial, temporal, event-level, and multi-hop reasoning. We intentionally targets implicit inferences that go beyond simple graph lookups, requiring a scene graph to preserve rich semantic nuances to support open-ended reasoning. During evaluation, an LLM answers the question using only the predicted scene graph as context; the resulting accuracy serves as a proxy for scene graph quality. By shifting from fixed query execution to holistic LLM reasoning, SGQA provides a more stringent measure of the scene graphs' utility.

\noindent{\textbf{(4) DSGG with Importance-aware Ranking.}} In addition to accuracy, the quality of scene graphs also depends on the \emph{importance} of the relations it contains. We therefore evaluate DSGG as a ranking problem based on relation importance, where we consider two criteria: \textit{informativeness} and \textit{diversity}:

\textit{Informativeness score.}
Prior work~\cite{sun2024chatgptgoodsearchinvestigating,mm_embed,ttev2} shows that LLMs can serve as effective zero-shot rerankers. We adopt a similar strategy to estimate how informative a triplet is with respect to a video clip. Given a triplet and the corresponding video segment, we prompt an MLLM with the query:
\texttt{"[video][triplet] Does the triplet represent an important action/event/visual component in the video? Yes or No."}
We then use the normalized probability of the \texttt{"Yes"} token as the informativeness score.

\textit{Diversity score.}
To encourage diversity among predicted relations, we measure how distinct each triplet is from previous ones, by encoding each triplet using a text encoder and compute the average cosine similarity between the new triplet and existing ones. Lower similarity indicates higher diversity.

The final triplet importance score is therefore a combination of informativeness and diversity scores. Given these scores, we propose to evaluate DSGG as a ranking task using Normalized Discounted Cumulative Gain (nDCG), a standard metric in information retrieval~\cite{mteb}:
\vspace{-5pt}
\[
\mathrm{DCG}_p = \sum_{i=1}^p{\frac{2^{rel_i}-1}{\log_2(i+1)}}, 
\quad
\mathrm{nDCG}_p = \frac{\mathrm{DCG}_p}{\mathrm{IDCG}_p}
\]
% \vspace{-5pt}
%
where $p$ denotes the rank position, $rel_i$ is the importance score of the $i$-th triplet, and $\mathrm{IDCG}_p$ is the ideal $\mathrm{DCG}$ obtained from the ground-truth ranking.

\noindent{\textbf{(5) From SGCls to \sgclsp.}} We propose a practical variant of the SGCls task. In the standard SGCls setting, models predict scene graphs given \emph{ground-truth} bounding boxes. However, such annotations are unavailable in real-world scenarios and therefore overlook the impact of detection errors on relation prediction. To better consider practical applications (VQA, retrieval, metadata storage, etc.) where grounding (\textit{i.e.} SGDet) is not required, we replace ground-truth bounding boxes with the model's predicted boxes during evaluation. We denote this setting as \sgclsp.

\begin{figure*}[t]
    \centering
    \includegraphics[width=\textwidth]{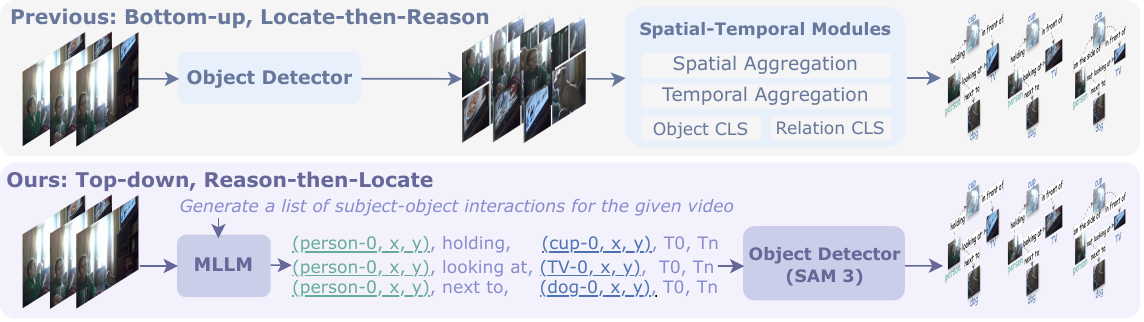}
    \vspace{-15pt}
    \caption{ Conventional DSGG pipelines (top) and ours (bottom). Existing methods typically follow a bottom-up approach: the detector extracts regional features, which are then processed by specialized modules. In contrast, our method adopts a top-down \emph{reason-then-locate} paradigm, where a MLLM first perform holistic reasoning, then grounded using a open-vocabulary detector.
    }
    \label{fig:diagram}
    \vspace{-10pt}
\end{figure*}

\subsection{Adapting MLLM for DSGG}

\noindent\textbf{Reason-then-Locate.} Traditional DSGG pipelines follow a bottom-up paradigm, where objects are first detected and relations are subsequently inferred. In contrast, we adopt a top-down \emph{reason-then-locate} strategy enabled by MLLMs. Specifically, the MLLM first performs holistic reasoning over the video to generate scene graph, and the grounding stage is performed afterward. Figure~\ref{fig:diagram} illustrates the conceptual difference between conventional DSGG pipelines and our approach.

\begin{wrapfigure}{r}{0.52\linewidth}
\centering
\vspace{-10pt}
\includegraphics[width=\linewidth]{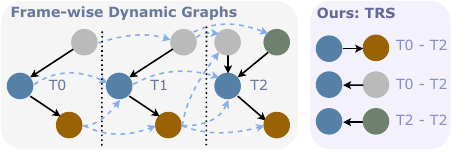}
\caption{\footnotesize Illustration of traditional frame-wise dynamic graphs (left) and the proposed Temporal Relation Set (TRS, right).}
\vspace{-18pt}
\label{fig:trs}
\end{wrapfigure}

\noindent\textbf{DSGG as Temporal Relation Set Prediction.} Existing DSGG methods typically model dynamic scene graphs as frame-wise graphs, where each frame contains a separate snapshot of object relations. Directly applying this formulation to MLLM-based generation leads to substantial redundancy, as the model must repeatedly generate identical triplets for every frame where the relation persists. To address this issue, we reformulate DSGG as a \emph{Temporal Relation Set (TRS)} prediction problem. Instead of frame-wise relations, each interaction is represented as a 5-ary, temporally grounded relation
\[
\langle \texttt{sub}, \texttt{rel}, \texttt{obj}, t_{\texttt{start}}, t_{\texttt{end}} \rangle .
\]
As illustrated in Figure~\ref{fig:trs}, this formulation eliminates duplicated predictions across frames and substantially improves both training and inference efficiency (Figure ~\ref{fig:trs_efficiency}). It also improves model accuracy, as excessive duplication during language-model finetuning can degrade generation quality.

\noindent\textbf{Grounding via Open-Vocabulary Detector.} Although recent MLLMs~\cite{qwen3vl,internvl3.5} possess basic grounding capabilities, they are not optimized for multi-object, temporally consistent video tracking. We therefore employ an off-the-shelf segmentation model. For each predicted temporal relation, the MLLM outputs the starting location ($x,y$) of the subject and object. We then utilize the referring expression capacity of frontier segmentation model (SAM 3~\citep{sam3}) to provide accurate grounding.

\noindent\textbf{Training MLLM-based DSGG with Importance-Aware Finetuning.} The autoregressive nature of MLLMs provides a simple mechanism to incorporate relation importance during generation. Specifically, during training we order the scene graph triplets according to their importance scores so that more important relations appear earlier in the sequence. The model is then finetuned using the standard negative log-likelihood objective:

\vspace{-10pt}
\[\mathcal{L} = -\sum^T_{t=1}\log \mathrm{P}(x_t | V, x_{<t}; \theta_{\texttt{MLLM}})\]
\vspace{-10pt}
\noindent where $V$ denotes the input video, $x_t$ denotes the $t$-th token.

\section{Experiments}

\subsection{Experimental Setup}

% \end{table}

\noindent\textbf{Datasets.}
Following prior DSGG works~\cite{sttran,tempura,oed}, we use the Action Genome (AG) dataset~\cite{ji2020actiongenome} as the primary benchmark, which contains \~{}10K videos annotated with 35 object and 25 relation categories. In addition, we evaluate our method on ImageNet-VidVRD~\cite{vidvrd} (VidVRD) and Panoptic Video Scene Graph (PVSG)~\cite{pvsg}, which extends DSGG to panoptic segmentation masks.

\noindent\textbf{Evaluation Metrics.}
For AG and VidVRD, we evaluate our method under the standard scene graph detection (SGDet) setting as well as our proposed \sgclsp\ protocol. Following previous works~\cite{sttran,dsg_detr,oed}, we consider both the \emph{With Constraint} and \emph{Without Constraint} settings and report recall@$k$ with $k \in \{10,20,50\}$. To assess the correctness of predicted relations, we additionally report precision@$k$ using the proposed \emph{MLLM-as-a-Judge} protocol with $k \in \{1,10,20\}$. For PVSG, the evaluation protocol is similar to the SGDet setting but replaces bounding boxes with segmentation masks. Following~\cite{uno,pvsg}, we report recall and mean recall@$k$ with $k \in \{20,50\}$.

\noindent\textbf{Implementation Details.}
We conduct experiments using two recent multimodal large language models: Qwen3VL~\cite{qwen3vl} and InternVL3.5~\cite{internvl3.5}. For visual grounding and tracking, we employ SAM~3~\cite{sam3}. The MLLMs are fine-tuned using the standard LoRA~\cite{lora} procedure with rank and scaling factor set to 32. For all datasets, models are trained for one epoch with a learning rate of $1\times10^{-5}$. Additional implementation details are provided in Appendix~\ref{sec:impl_details}.

\subsection{Main Results}

\begin{table*}
    \centering
    % === Left Table ===
    \begin{minipage}[b]{0.50\textwidth}
        \centering
        \caption{\sgclsp performance under With/No Constraints settings on AG. Best/2nd best are highlighted in \textbf{bold}/\underline{underlined}. Deltas are computed based on the SoTA baseline.\protect\footnotemark}
        \vspace{2pt}
        \setlength{\tabcolsep}{3.5pt}
        \resizebox{\linewidth}{!}{
            \begin{tabular}{lccc c ccc}
            \toprule
            & \multicolumn{3}{c}{\textbf{With Constraint}} & & \multicolumn{3}{c}{\textbf{No Constraint}} \\
            \cmidrule(r){2-4} \cmidrule(l){6-8}
            Method & R@10 & R@20 & R@50 & & R@10 & R@20 & R@50 \\
            \midrule
            STTran  & 30.2 & 37.6 & 44.2 && 31.1 & 45.7 & 62.1 \\
            TEMPURA & 33.6 & 36.4 & 42.8 && 37.8 & 48.5 & 59.7 \\
            TRACE   & 20.3 & 21.7 & 21.8 && 26.4 & 43.1 & 52.6 \\
            OED     & 41.2 & 52.3 & 63.6 && 44.3 & 55.9 & 66.8 \\
            \midrule
            \rowcolor{lblue}
            \textbf{Ours} (InternVL-3.5) & \underline{43.5}~\inc{2.3} & \underline{56.6}~\inc{4.3} & \underline{63.8}~\inc{0.2} && 44.2~\dec{0.1} & \underline{56.8}~\inc{0.9} & \underline{67.7}~\inc{0.9} \\
            \rowcolor{lblue}
            \textbf{Ours} (Qwen3VL) & \textbf{47.3}~\inc{6.1} & \textbf{61.1}~\inc{8.8} & \textbf{71.7}~\inc{8.1} && \textbf{46.6}~\inc{2.3} & \textbf{60.3}~\inc{4.4} & \textbf{69.5}~\inc{2.7} \\
            \bottomrule
            \end{tabular}
        }
        \label{tab:sgclsp}
        \vspace{-10pt}
    \end{minipage}
    \hfill % Pushes the two tables apart
    % === Right Table ===
    \begin{minipage}[b]{0.48\textwidth}
        \centering
        \caption{\sgclsp and SGDet results on VidVRD dataset. Deltas are computed against the strongest baseline per task.}
        % \vspace{2pt}
        \setlength{\tabcolsep}{6pt}
        \resizebox{\linewidth}{!}{
            \begin{tabular}{lccc c ccc}
            \toprule
            & \multicolumn{3}{c}{\textbf{Recall}} & & \multicolumn{3}{c}{\textbf{Precision}} \\
            \cmidrule(r){2-4} \cmidrule(l){6-8}
            Method & @10 & @20 & @50 & & @1 & @10 & @20 \\
            \midrule
            \multicolumn{8}{c}{\textbf{\sgclsp}} \\
            VrdONE & 19.1 & 27.8 & 43.0 && \textbf{80.5} & 46.2 & 33.6 \\
            OED\protect\footnotemark & 24.4 & 31.4 & 55.2 && 61.4 & 49.7 & 38.2 \\
            \rowcolor{lblue}
            \textbf{Ours} (InternVL-3.5) & 29.2\,\inc{4.8} & 38.4\,\inc{7.0} & 61.5\,\inc{6.3} && 69.1\,\dec{11.4} & 61.3\,\inc{11.6} & 53.4\,\inc{15.2} \\
            \rowcolor{lblue}
            \textbf{Ours} (Qwen3VL) & \textbf{32.1}\,\inc{7.7} & \textbf{40.3}\,\inc{8.9} & \textbf{65.2}\,\inc{10.0} && 72.3\,\dec{8.2} & \textbf{66.2}\,\inc{16.5} & \textbf{55.3}\,\inc{17.1} \\
            \midrule
            \multicolumn{8}{c}{\textbf{SGDet}} \\
            VrdONE & 10.1 & 13.6 & 18.2 && 53.0 & 24.5 & 16.4 \\
            OED & 19.4 & 24.3 & 22.8 && 37.5 & 31.3 & 24.9 \\
            \rowcolor{lblue}
            \textbf{Ours} (InternVL-3.5) & 26.6\,\inc{7.2} & 33.7\,\inc{9.4} & 41.3\,\inc{18.5} && 54.8\,\inc{1.8} & 45.6\,\inc{14.2} & 41.2\,\inc{16.3} \\
            \rowcolor{lblue}
            \textbf{Ours} (Qwen3VL) & \textbf{31.5}\,\inc{12.1} & \textbf{38.2}\,\inc{13.9} & \textbf{48.6}\,\inc{25.8} && \textbf{56.1}\,\inc{3.1} & \textbf{51.4}\,\inc{20.2} & \textbf{45.2}\,\inc{20.3} \\
            \bottomrule
            \end{tabular}
        }
        \label{tab:vidvrd}
    \vspace{-10pt}
    \end{minipage}
\end{table*}

\footnotetext{Note that since we need to re-evaluate baselines for \sgclsp, we only report models which provide codebase and for which we can reproduce the reported results.}

% \noindent\textbf{{Results on Action Genome.}} Table~\ref{tab:sgdet_ag} and~\ref{tab:sgclsp} present the comparison with prior DSGG methods on the Action Genome dataset. Our MLLM-based approach consistently outperforms existing models across both the standard SGDet setting and the proposed \sgclsp protocol. 
% Under the SGDet setting (Table~\ref{tab:sgdet_ag}), our method achieves substantial improvements over the previous state-of-the-art (UNO). In particular, ours (Qwen3VL) achieves $49.6/57.9/62.2$ recall at R$@10/20/50$ under the \emph{With Constraint} setting, improving over UNO by $+10.3$, $+12.7$, and $+8.4$, respectively. Similar gains are observed under the \emph{No Constraint} setting, where our method improves $R@50$ from $57.1$ to $61.7$. 

\noindent\textbf{Results on Action Genome (AG).} 
As shown in Table~\ref{tab:sgdet_ag} and~\ref{tab:sgclsp}, our MLLM-based approach consistently outperforms existing DSGG models across both SGDet and \sgclsp. Under the SGDet \emph{With Constraint} setting, Ours (Qwen3VL) achieves $49.6/57.9/62.2$ at $R@10/20/50$, surpassing the SOTA (UNO) by $+10.3$, $+12.7$, and $+8.4$, respectively. Similar gains occur under \emph{No Constraint}, with $R@50$ reaching $61.7$ ($+4.6$ over UNO). Furthermore, Table~\ref{tab:precision_ag} shows significant precision gains; e.g., under SGDet, $P@10$ improves from $27.2$ (OED) to $48.4$. These results indicate that our method generates fewer redundant relations, yielding a superior precision-recall trade-off (Figure~\ref{fig:pre-recall}).

\begin{wraptable}{r}{0.6\linewidth}
\vspace{-12pt}
\centering
\caption{ Results under With/No Constraints settings for SGDet on AG. Deltas are computed based on the SoTA baseline (UNO).}
\vspace{-5pt}
\resizebox{\linewidth}{!}{
\begin{tabular}{lccc c ccc}
\toprule
& \multicolumn{3}{c}{\textbf{With Constraint}} 
& 
& \multicolumn{3}{c}{\textbf{No Constraints}} \\
\cmidrule(r){2-4} \cmidrule(l){6-8}
Method 
& R@10 & R@20 & R@50 
& 
& R@10 & R@20 & R@50 \\
\midrule
VRD~\cite{vrd} & 19.2 & 24.5 & 26.0 && 19.1 & 28.8 & 40.5 \\
GPS-Net~\cite{gpsnet} & 24.7 & 33.1 & 35.1 && 24.4 & 35.7 & 47.3 \\
STTran~\cite{sttran} & 25.2 & 34.1 & 37.0 && 24.6 & 36.2 & 48.8 \\
TRACE~\cite{trace} & 13.9 & 14.5 & 14.5 && 26.5 & 35.6 & 45.3 \\
TEMPURA~\cite{tempura} & 28.1 & 33.4 & 34.9 && 29.8 & 38.1 & 46.4 \\
TR$^2$~\cite{tr2} & 26.8 & 35.5 & 38.3 && 27.8 & 39.2 & 50.0 \\
APT~\cite{apt} & 26.3 & 36.1 & 38.3 && 25.7 & 37.9 & 50.1 \\
RelTR~\cite{reltr} & 19.7 & 23.4 & 25.9 && 20.9 & 24.6 & 28.2 \\
DSG-DETR~\cite{dsg_detr} & 30.3 & 34.8 & 36.1 && 32.1 & 40.9 & 48.3 \\
TPT~\cite{tpt} & - & - & - && 32.0 & 39.6 & 51.5 \\
OED~\cite{oed} & 33.5 & 40.9 & 48.9 && 35.3 & 44.0 & 51.8 \\
UNO~\cite{uno} & 39.3 & 45.2 & 53.8
&& 40.8 & 49.7 & 57.1 \\
ARN~\cite{arn} & 35.1 & 41.8 & 47.2 && 37.6 & 46.8 & 54.1 \\
RS-Net~\cite{rsnet} & 30.5 & 35.0 & 36.3 && - & - & - \\
\midrule
\rowcolor{lblue}
\textbf{Ours} (InternVL-3.5) & \underline{43.1}~\inc{3.8} & \underline{55.8}~\inc{10.6} & \underline{59.3}~\inc{5.5}
&& \underline{42.4}~\inc{1.6} & \underline{53.2}~\inc{3.5} & \underline{58.8}~\inc{1.7} \\
\rowcolor{lblue}
\textbf{Ours} (Qwen3VL) & \textbf{49.6}~\inc{10.3} & \textbf{57.9}~\inc{12.7} & \textbf{62.2}~\inc{8.4}
&& \textbf{47.3}~\inc{6.5} & \textbf{56.2}~\inc{6.4} & \textbf{61.7}~\inc{4.6} \\
\bottomrule
\end{tabular}
}
\label{tab:sgdet_ag}
\vspace{-10pt}
\end{wraptable}

% Table~\ref{tab:precision_ag} further reports precision under both SGDet and \sgclsp under AG. Our method achieves significantly higher precision than existing DSGG models. For example, under SGDet our approach improves $P@10$ from $27.2$ (OED) to $48.4$. These improvements suggest that our MLLM-based DSGG approach generates substantially fewer noisy or redundant relations, thereby achieving notably better precision-recall trade-off (Figure~\ref{fig:pre-recall}).

% \noindent\textbf{{Results on VidVRD.}} Table~\ref{tab:vidvrd} reports the results on the VidVRD dataset under SGDet and \sgclsp\ settings. Our approach consistently outperforms previous methods across both recall and precision metrics. Under the \sgclsp\ setting, Ours (Qwen3VL) achieves $32.1/40.3/65.2$ recall at $R@10/20/50$, improving over the strongest baseline (OED) by $+7.7$, $+8.9$, and $+10.0$, respectively. Similarly, under SGDet our method achieves substantial gains, reaching $48.6$ at $R@50$.

\noindent\textbf{Results on VidVRD.} 
Table~\ref{tab:vidvrd} shows that our approach maintains its performance edge on VidVRD. Under \sgclsp, Ours (Qwen3VL) achieves $32.1/40.3/65.2$ at $R@10/20/50$, outperforming the strongest baseline (OED) by $+7.7$, $+8.9$, and $+10.0$, respectively. In SGDet, our method achieves an $R@50$ of $48.6$, consistently leading in both recall and precision across all evaluated metrics.

\noindent\textbf{{Results on PVSG.}}
Table~\ref{tab:pvsg} reports the results on the PVSG benchmark under different vIoU thresholds. Our method consistently outperforms prior approaches across both recall and mean recall metrics. In particular, Ours (Qwen3VL) achieves $17.8/17.1$ R/mR@20 and $19.2/16.3$ R/mR@50 under $\text{vIoU}=0.5$, improving over the SOTA (UNO) by large margins. These results demonstrate that the proposed MLLM-based framework generalizes well to panoptic scene graph generation.

\footnotetext{While the original OED~\cite{oed} does not test on VidVRD, we run training using OED's codebase with the default configurations from AG for VidVRD.}

% \begin{figure}[H]
%     \vspace{-10pt}
%     \centering
%     \includegraphics[width=\linewidth]{images/takeaway1.pdf}
%     \label{fig:placeholder}
%     \vspace{-10pt}
% \end{figure}

\begin{table*}
\centering
% \resizebox{\linewidth}{!}{
\begin{minipage}[b]{0.49\textwidth}
\centering
\captionof{table}{Effect of Importance-Aware Finetuning (IAF) on ranking performance. Deltas are computed \textit{w.r.t.} w/o IAF.}
\vspace{-5pt}
\resizebox{\linewidth}{!}{
\begin{tabular}{lccc}
\toprule
Model & w/ IAF & nDCG@5 & nDCG@10 \\
\midrule
TEMPURA        & - & 27.1 & 31.6 \\
DSG-DETR       & - & 29.4 & 32.8 \\
OED            & - & 32.3 & 35.2 \\
\midrule
\textbf{Ours} (InternVL-3.5)   & \xmark & 35.7 & 38.4 \\
% \rowcolor{lightergray}
% \rowcolor{lblue}
                               & \cmark & 43.0~\inc{7.3} & 45.2\,\inc{6.8} \\
\textbf{Ours} (Qwen3VL)        & \xmark & 37.9 & 41.6 \\
% \rowcolor{lightergray}
% \rowcolor{lblue}
                               & \cmark & 44.3~\inc{6.4} & 47.6\,\inc{5.9} \\
\bottomrule
\end{tabular}
}
\label{tab:iaf}

\end{minipage}
\vspace{-5pt}
\hfill
\begin{minipage}[b]{0.49\textwidth}
\centering
\captionof{table}{Performance on SGRet and SGQA. Best results are \textbf{bold}, second best are \underline{underlined}.}
\vspace{-5pt}
\resizebox{\linewidth}{!}{
\begin{tabular}{lcc c c}
\toprule
& \multicolumn{2}{c}{\textbf{SGRet}} 
& 
& \textbf{SGQA} \\
\cmidrule(r){2-3} \cmidrule(l){5-5}
Method & R@5 & R@10 
& 
& Acc (\%) \\
\midrule
GT           & {68.4} & {80.2} && {63.8} \\
\midrule
STTran       & 8.41  & 14.3 && 25.5 \\
TEMPURA      & 14.3 & 20.6 && 31.6 \\
OED          & 16.3 & 22.3 && 36.2 \\
\midrule
\rowcolor{lblue}
\textbf{Ours} (InternVL3.5)  & \underline{35.9}~\inc{19.5} & \underline{44.5}~\inc{22.1} && \underline{47.8}~\inc{11.75} \\
\rowcolor{lblue}
\textbf{Ours} (Qwen3VL)     & \textbf{36.8}~\inc{20.44} & \textbf{46.4}~\inc{24.1} && \textbf{48.9}~\inc{12.8} \\
\bottomrule
\end{tabular}
}
\label{tab:sgret_sgqa}
\end{minipage}
\vspace{-5pt}
\end{table*}

\subsection{Ablations}

\noindent\textbf{Effect of Importance-Aware Finetuning (IAF).} Table~\ref{tab:iaf} evaluates the impact of our proposed IAF on relation ranking. Without IAF, both prior DSGG methods and our MLLM-based models show limited awareness of triplet importance, yielding lower ranking quality; specifically, our baseline results are comparable to bottom-up methods like OED. This confirms that standard MLLM finetuning does not inherently prioritize informative triplets. Incorporating IAF yields notable gains: the InternVL3.5-based model improves by $+7.3$ ($43.0$) in nDCG@5 and $+6.8$ ($45.2$) in nDCG@10, while the Qwen3VL-based model reaches $44.3$ and $47.6$ in nDCG@5/10, respectively. These improvements demonstrate that explicitly modeling importance during finetuning enables more informative and diverse relation generation, as qualitatively validated in Figure~\ref{fig:uninformative}.

\begin{wraptable}{r}{0.55\linewidth}
\centering
\vspace{-12pt}
\caption{Performance comparison of PVSG under different vIoUs.}
\vspace{-5pt}
\setlength{\tabcolsep}{4pt}
\resizebox{\linewidth}{!}{%
\begin{tabular}{lcc c cc}
\toprule
& \multicolumn{2}{c}{vIoU = 0.5} && \multicolumn{2}{c}{vIoU = 0.1} \\
\cmidrule(r){2-3}\cmidrule(l){5-6}
Method & R/mR@20 & R/mR@50 && R/mR@20 & R/mR@50 \\
\midrule
VPS~\cite{pvsg}   & 0.42 / 0.61 & 0.73 / 0.76 && 6.50 / 5.75 & 9.64 / 8.25 \\
IPS+T~\cite{pvsg} & 3.88 / 2.81 & 5.66 / 4.12 && 9.01 / 6.69 & 14.9 / 11.3 \\
VISA~\cite{visa}  & \ \ - \ \  / 8.82    & \ \ - \ \  / 8.94    && -           & - \\
MCL~\cite{mcl}   & 0.84 / 0.98 & 1.26 / 1.22 && 8.18 / 8.00 & 12.9 / 11.5 \\
UNO~\cite{uno}   & 9.44 / 8.25 & 10.8 / 9.72 && 17.5 / 14.8 & 23.8 / 18.3 \\
\midrule
\rowcolor{lblue}
\textbf{(Ours)} InternVL3.5 & \underline{13.4} / \underline{12.2} & \underline{15.8} / \underline{14.6} && \underline{21.1} / \underline{20.6} & \underline{26.5} / \underline{25.4} \\
\rowcolor{lblue}
\textbf{(Ours)} Qwen3-VL    & \textbf{17.8} / \textbf{17.1} & \textbf{19.2} / \textbf{16.3} && \textbf{26.5} / \textbf{23.2} & \textbf{31.3} / \textbf{30.8} \\
\bottomrule
\end{tabular}%
}
\vspace{-12pt}
\label{tab:pvsg}
\end{wraptable}

% By contrast, incorporating IAF leads to notable improvements in ranking performance. After applying IAF, the InternVL3.5-based model improves from $35.7$ to $43.0$ in nDCG@5 and from $38.4$ to $45.2$ in nDCG@10, while the Qwen3VL-based model improves from $37.9$ to $44.3$ and from $41.6$ to $47.6$, respectively. These large gains indicate that explicitly modeling triplet importance during finetuning enables the model to generate relations in a more informative and diverse order, which can also be visually validated from Figure~\ref{fig:uninformative}.

\noindent\textbf{{Downstream Task Evaluation (SGRet and SGQA).}}
Table~\ref{tab:sgret_sgqa} evaluates the quality of generated scene graphs through two downstream tasks: SGRet and SGQA. Our MLLM-based approach substantially outperforms prior DSGG methods on both tasks. Notable improvements are observed on SGQA, where our method reaches $48.9\%$ accuracy compared to $36.2\%$ from OED. Notably, the improvements are much larger than those observed on traditional recall metrics, suggesting that existing DSGG models often generate relations that match annotations but fail to capture the key semantics required for downstream reasoning. In contrast, our approach produces more informative and coherent scene graphs, which directly translate into stronger retrieval and question answering performance.

\begin{wraptable}{r}{0.58\linewidth}
% \begin{table*}
\centering
\vspace{-10pt}
\caption{Precision on SGDet and \sgclsp under AG. Deltas are computed based on the SoTA baseline (OED).}
\vspace{-5pt}
\resizebox{\linewidth}{!}{
\begin{tabular}{lccc c ccc}
\toprule
& \multicolumn{3}{c}{SGDet} 
& 
& \multicolumn{3}{c}{\sgclsp} \\
\cmidrule(r){2-4} \cmidrule(l){6-8}
Method 
& P@1 & P@10 & P@20
& 
& P@1 & P@10 & P@20 \\
\midrule
STTran   & 26.3 & 17.6 & 13.1 && 32.7 & 22.6 & 16.9 \\
TRACE    & 27.5 & 15.3 & 12.7 && 26.5 & 18.7 & 13.7 \\
TEMPURA  & 38.1 & 21.4 & 13.7 && 49.1 & 27.3 & 18.0 \\
DSG-DETR & 37.4 & 22.3 & 14.4 && 51.3 & 30.6 & 20.2 \\
OED      & 43.6 & 27.2 & 17.4 && 54.1 & 35.2 & 22.8 \\
\midrule
\rowcolor{lblue}
\textbf{Ours} (InternVL-3.5)
         & \underline{48.2}~\inc{4.6} & \underline{42.9}~\inc{15.7} & \underline{30.4}~\inc{13.0}
         && \underline{58.3}~\inc{4.2} & \underline{50.9}~\inc{15.7} & \underline{36.4}~\inc{13.6} \\
\rowcolor{lblue}
\textbf{Ours} (Qwen3VL)
         & \textbf{53.0}~\inc{9.4} & \textbf{48.4}~\inc{21.2} & \textbf{35.9}~\inc{18.5} 
         && \textbf{59.1}~\inc{5.0} & \textbf{53.1}~\inc{18.0} & \textbf{38.2}~\inc{15.4} \\
\bottomrule
\end{tabular}
}
\vspace{-10pt}
\label{tab:precision_ag}
\end{wraptable}

\begin{wrapfigure}{r}{0.37\linewidth}
\centering
\vspace{-14pt}
\includegraphics[width=\linewidth]{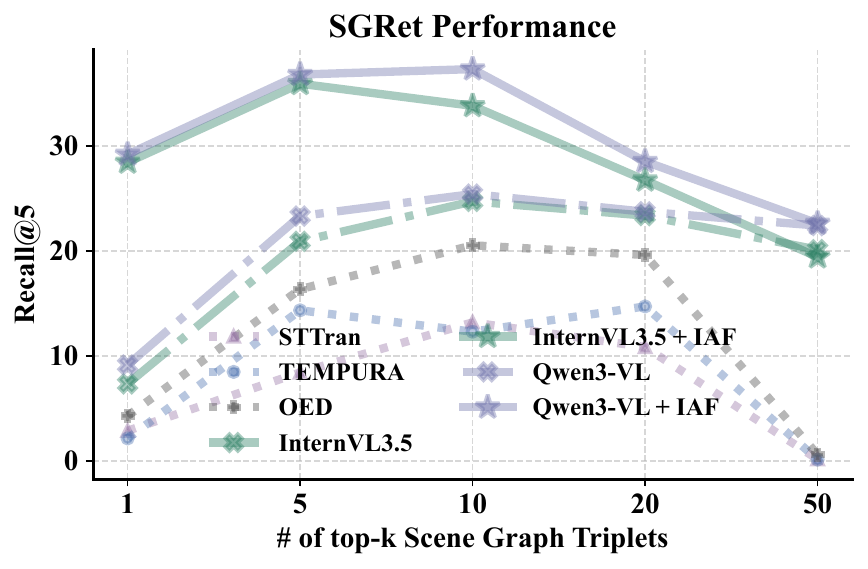}
\vspace{-15pt}
\captionof{figure}{SGRet's recall@5 performance on Action Genome dataset as a function of the number of top-$k$ triplets used during retrieval.}
\label{fig:sgret_k}
\vspace{-10pt}
\end{wrapfigure}

Figure~\ref{fig:sgret_k} further analyzes SGRet performance as a function of the number of top-$k$ scene graph triplets used during retrieval. Overall, MLLM-based approaches consistently outperform traditional DSGG models across different values of $k$. Particularly, existing methods exhibit an obvious concave trend: retrieval performance is poor when only the top-$1$ triplet is used and also deteriorates when a large number of triplets ($k=50$) are included. The weak performance at $k=1$ suggests that the highest-confidence triplets produced by these models are often not the most informative ones. Meanwhile, the drop at larger $k$ reflects the accumulation of noisy and redundant relations, which lowers retrieval precision.

In contrast, our approach maintains strong performance across a wide range of $k$. When IAF is disabled, our model exhibits a similar degradation at $k=1$, confirming that standard training objectives lack awareness of triplet importance. After incorporating IAF, however, the model learns to prioritize informative relations, leading to strong retrieval performance even when only a few triplets are used. Moreover, our method remains robust when $k$ increases, as MLLMs naturally ''know`` when to stop generation. This behavior contrasts with conventional DSGG models, which tend to over-generate relations and therefore suffer significant precision loss when larger triplet sets are used.

\begin{figure}
    % \vspace{-15pt}
    \centering
    \includegraphics[width=\columnwidth]{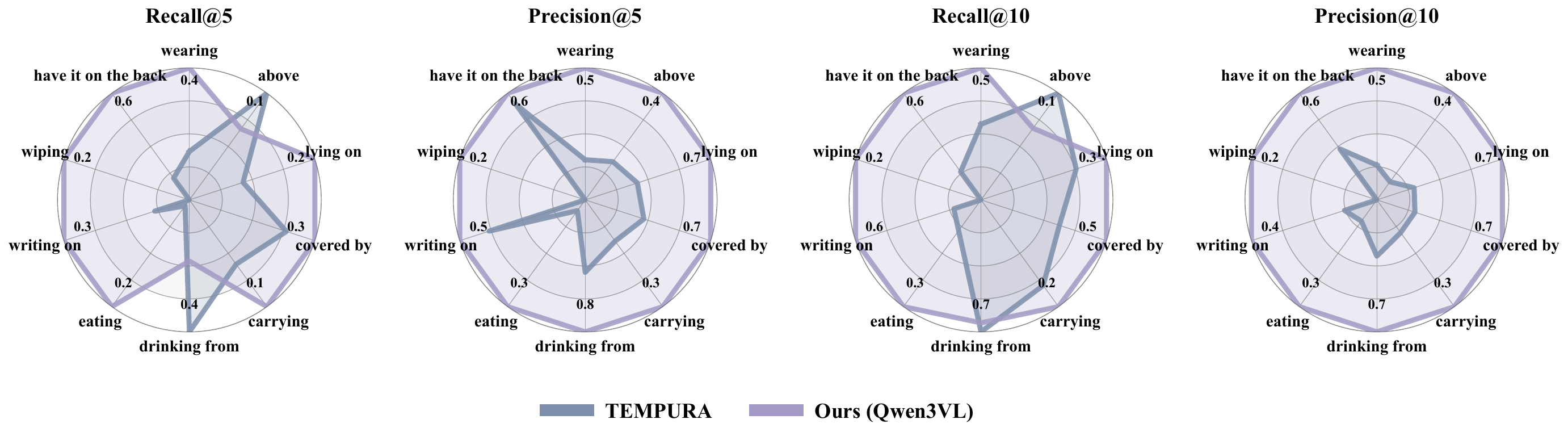}
    \vspace{-20pt}
    \caption{ Per relation category Recall/Precision performance between our MLLM-based method  (purple) and TEMPURA (blue) on Action Genome dataset, under the \sgclsp setting. We show the 15 \textbf{\textit{least}}-frequent relation categories.}
    \label{fig:class_wise}
    \vspace{-4pt}
\end{figure}

\begin{figure}
\centering

\begin{minipage}[t]{0.45\linewidth}
\vspace{-5pt}
\centering
\includegraphics[width=\linewidth]{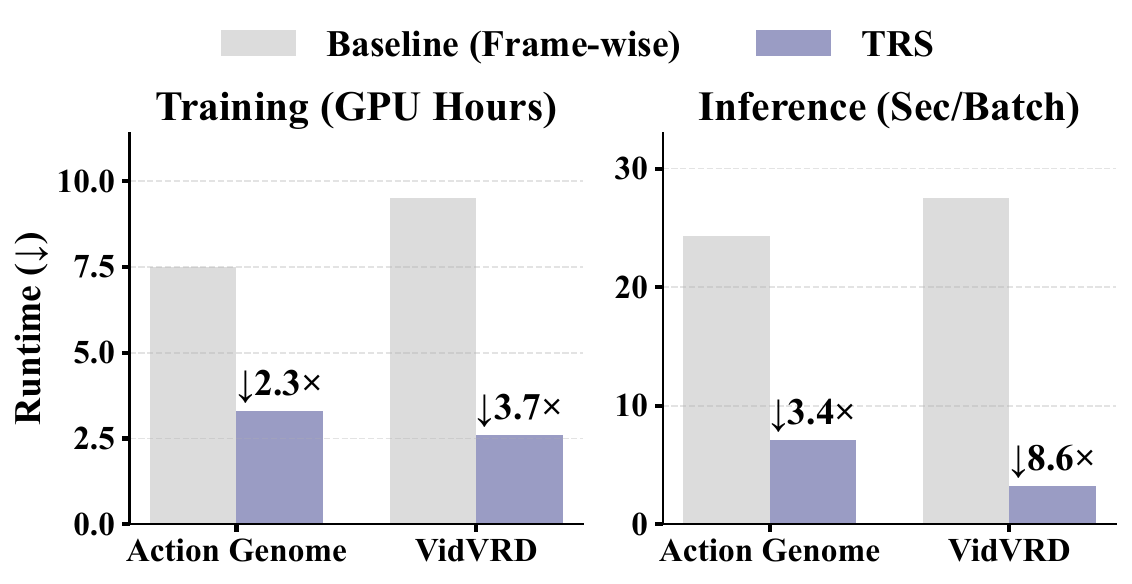}
\vspace{-18pt}
\caption{Training/Inference speed w/ or w/o TRS.}
\label{fig:trs_efficiency}
\end{minipage}
\vspace{-10pt}
\hfill
\begin{minipage}[t]{0.52\linewidth}
\vspace{0pt}
\centering
\captionof{table}{Performance comparison on Action Genome with or without adopting the proposed Temporal Relation Set (TRS) representation of dynamic scene graphs.}
% \vspace{5pt}
\resizebox{\linewidth}{!}{%
\begin{tabular}{lc cc c cc}
\toprule
\multirow{2}{*}{Backbone} & \multirow{2}{*}{w/ TRS} & \multicolumn{2}{c}{\textbf{Action Genome}} 
& & \multicolumn{2}{c}{\textbf{VidVRD}} \\
\cmidrule(r){3-4} \cmidrule(l){6-7}
 &  & R@10 & P@1 && R@10 & P@1 \\
\midrule
InternVL3.5 & \xmark & 42.9 & 57.6 && 28.4 & 70.3 \\
% \rowcolor{lightergray}
% \rowcolor{lblue}
            & \cmark & {44.2}\,\inc{1.3} 
                     & {58.3}\,\inc{0.7} 
                     && {29.2}\,\inc{0.8} 
                     & 69.1\,\dec{-1.2} \\
\midrule
Qwen3-VL    & \xmark & 45.3 & 58.3 && 30.4 & 73.1 \\
% \rowcolor{lightergray}
% \rowcolor{lblue}
            & \cmark & \textbf{46.6}\,\inc{1.3} 
                     & \textbf{59.1}\,\inc{0.8} 
                     && \textbf{32.1}\,\inc{1.7} 
                     & \textbf{73.9}\,\inc{0.8} \\
\bottomrule
\end{tabular}
}
\label{tab:trs_abl}
\end{minipage}
\vspace{-10pt}
\end{figure}

% \noindent\textbf{{Effect of Temporal Relation Set.}}
% Figure~\ref{fig:trs_efficiency} compares the training and inference efficiency between the naive frame-wise formulation of DSGG and the proposed Temporal Relation Set (TRS) representation. The naive formulation requires the model to repeatedly generate identical relations for every frame in which the interaction occurs, resulting in substantial redundancy during both training and inference. By representing relations as temporally grounded tuples, TRS eliminates these duplicated predictions and significantly reduces the generation length. As a result, TRS leads to notable efficiency gains across datasets. In particular, training becomes $2.3\times$ and $3.7\times$ faster on Action Genome and VidVRD, respectively, while inference latency is reduced by $3.4\times$ and $8.6\times$. These results demonstrate that the proposed TRS formulation substantially improves the computational efficiency of MLLM-based DSGG.

\noindent\textbf{Impact of Temporal Relation Set (TRS).} Figure~\ref{fig:trs_efficiency} and Table~\ref{tab:trs_abl} evaluate the efficiency and accuracy gains of our TRS representation over the redundant, frame-wise DSGG baseline. By representing relations as temporally grounded tuples, TRS eliminates duplicated predictions across frames, drastically reducing sequence length. Consequently, training speed increases by $2.3\times$ on AG, while inference latency drops by up to $8.6\times$. Beyond efficiency, TRS consistently improves performance; for instance, $R@10$ increases by up to $+1.3$ and $+1.7$ on Action Genome and VidVRD, respectively. These results suggest that removing frame-wise redundancy provides a cleaner supervision signal, allowing MLLMs to achieve higher accuracy with significantly lower computational overhead.

% Table~\ref{tab:trs_abl} further evaluates the impact of TRS on prediction accuracy. Despite primarily targeting efficiency, TRS also consistently improves performance across both backbones and datasets. On Action Genome, adopting TRS improves recall and precision for both InternVL3.5 and Qwen3-VL models, with gains of up to $+1.3$ in $R@10$ and $+0.8$ in $P@1$. Similar improvements are observed on VidVRD, where TRS increases $R@10$ by $+0.8$ and $+1.7$ for the two backbones, respectively. These results suggest that removing duplicated frame-wise supervision not only accelerates training but also leads to cleaner learning signals for MLLMs, resulting in more accurate relation prediction.

\noindent\textbf{{Long-tailed prediction with MLLM-based DSGG.}}
Figure~\ref{fig:class_wise} compares per-category accuracy between our MLLM-based approach and TEMPURA, a method specifically designed to mitigate long-tail issues. Nevertheless, TEMPURA still exhibits highly uneven behavior, achieving moderate performance on a few relations but failing on many others. In contrast, our MLLM-based model maintains consistent performance across most relation categories, including rare predicates such as \textit{wiping}, \textit{writing on}, and \textit{have it on the back}. This behavior suggests that MLLMs benefit from their large-scale pretraining and broader visual-semantic knowledge, allowing them to generalize beyond the frequency biases present in DSGG datasets. As a result, our approach acts as a more balanced scene graph generator, improving both coverage and reliability for long-tail relations.

\subsection{Further Analysis}
\label{sec:analysis}

\begin{table}[t]
\centering
\begin{minipage}[b]{0.45\linewidth}
\centering
\caption{\sgclsp $P@10$ on AG under three MLLM judges: \mbox{Qwen3-VL-30B-A3B} (default), Gemini 3.5 \mbox{Flash-Lite}, and Gemma 4 \mbox{26B-A4B}. Std: standard deviation across judges.}
\vspace{2pt}
\resizebox{\linewidth}{!}{
\begin{tabular}{lcccc}
\toprule
Method & Qwen3-VL & Gemini & Gemma & Std \\
\midrule
DSG-DETR & 30.6 & 29.9 & 31.2 & 0.65 \\
OED      & 35.2 & 33.6 & 33.4 & 0.99 \\
\midrule
\rowcolor{lblue}
\textbf{Ours} (InternVL-3.5) & 50.9 & 50.1 & 51.7 & 0.80 \\
\rowcolor{lblue}
\textbf{Ours} (Qwen3VL)      & 53.1 & 54.3 & 52.5 & 0.92 \\
\bottomrule
\end{tabular}
}
\label{tab:judge}
\end{minipage}
\hfill
\begin{minipage}[b]{0.52\linewidth}
\centering
\caption{Error analysis. \textbf{FP}: false positives due to wrong temporal intervals (Temp.), wrong relations (Rel.), or other errors. \textbf{FN}: missed relations with below-mean informativeness (Low-Info.) or duration (Short), which can overlap.}
\vspace{2pt}
\resizebox{\linewidth}{!}{
\begin{tabular}{lccc c cc}
\toprule
& \multicolumn{3}{c}{\textbf{FP (\%)}} && \multicolumn{2}{c}{\textbf{FN (\%)}} \\
\cmidrule(r){2-4} \cmidrule(l){6-7}
Method & Temp. & Rel. & Other && Low-Info. & Short \\
\midrule
OED & 44.1 & 38.7 & 17.2 && 48.2 & 56.8 \\
\midrule
\rowcolor{lblue}
\textbf{Ours} (InternVL-3.5) & 74.6 & 23.1 & 2.3 && 83.6 & 67.3 \\
\rowcolor{lblue}
\textbf{Ours} (Qwen3VL)      & 69.5 & 22.3 & 8.2 && 80.4 & 60.1 \\
\bottomrule
\end{tabular}
}
\label{tab:error}
\end{minipage}

\vspace{8pt}
\begin{minipage}[b]{0.49\linewidth}
\centering
\caption{Comparison with zero-shot (ZS) proprietary MLLMs on AG (\sgclsp, No Constraint).}
\vspace{2pt}
\resizebox{\linewidth}{!}{
\begin{tabular}{lccc c ccc c cc}
\toprule
& \multicolumn{3}{c}{\textbf{Precision}} && \multicolumn{3}{c}{\textbf{Recall}} && \multicolumn{2}{c}{\textbf{nDCG}} \\
\cmidrule(r){2-4} \cmidrule(lr){6-8} \cmidrule(l){10-11}
Method & @1 & @10 & @20 && @10 & @20 & @50 && @5 & @10 \\
\midrule
OED                        & 54.1 & 35.2 & 22.8 && \underline{44.3} & \underline{55.9} & \underline{66.8} && \underline{32.3} & \underline{35.2} \\
GPT-5.6 Luna (ZS)          & \underline{63.9} & \underline{55.4} & \underline{38.2} && 24.5 & 45.2 & 59.1 && 16.6 & 18.4 \\
Gemini 3.5 Flash-Lite (ZS) & \textbf{68.4} & \textbf{57.6} & \textbf{40.7} && 27.3 & 43.7 & 56.4 && 17.9 & 18.2 \\
\midrule
\rowcolor{lblue}
\textbf{Ours} (Qwen3VL)    & 59.1 & 53.1 & \underline{38.2} && \textbf{46.6} & \textbf{60.3} & \textbf{69.5} && \textbf{44.3} & \textbf{47.6} \\
\bottomrule
\end{tabular}
}
\label{tab:zs_frontier}
\end{minipage}
\hfill
\begin{minipage}[b]{0.49\linewidth}
\centering
\caption{Open-vocabulary results on VidVRD (Novel split) under the OpenVidVRD protocol~\cite{openvidvrd}, from which baseline results are taken. 50\%/100\%: fraction of training data.}
\vspace{2pt}
\resizebox{\linewidth}{!}{
\begin{tabular}{lccc c ccc}
\toprule
& \multicolumn{3}{c}{\textbf{SGDet}} && \multicolumn{3}{c}{\textbf{SGCls}} \\
\cmidrule(r){2-4} \cmidrule(l){6-8}
Method & mAP & R@50 & R@100 && mAP & R@50 & R@100 \\
\midrule
ALPro~\cite{alpro}           & 0.98 & 2.79 & 4.33 && 3.69 & 7.27 & 8.92 \\
VidVRD-II~\cite{vidvrdii}    & 3.11 & 7.93 & 11.38 && 5.70 & 13.22 & 18.34 \\
RePro~\cite{repro}           & 5.87 & 12.75 & 16.23 && 10.32 & 19.17 & 25.28 \\
MMP~\cite{mmp}               & 12.15 & 13.75 & 15.21 && 17.57 & 21.98 & 28.43 \\
OpenVidVRD~\cite{openvidvrd} & 17.95 & 17.69 & 19.01 && 18.25 & 28.98 & 31.85 \\
\midrule
\rowcolor{lblue}
\textbf{Ours} (Qwen3VL, 50\%)  & \textbf{32.13} & \textbf{33.85} & \textbf{33.87} && \textbf{37.63} & \textbf{40.29} & \textbf{40.32} \\
\rowcolor{lblue}
\textbf{Ours} (Qwen3VL, 100\%) & \underline{29.54} & \underline{32.66} & \underline{32.69} && \underline{32.41} & \underline{36.83} & \underline{36.87} \\
\bottomrule
\end{tabular}
}
\label{tab:openvoc}
\end{minipage}
\vspace{-10pt}
\end{table}

\noindent\textbf{Robustness of MLLM-as-a-Judge.} As our default judge (Qwen3-VL-30B-A3B) belongs to the same model family as one of our generators, we examine whether precision depends on the judge. Table~\ref{tab:judge} re-evaluates \sgclsp $P@10$ on AG with two judges from independent model families, Gemini 3.5 Flash-Lite~\cite{gemini3_5flashlite} and Gemma 4 26B-A4B~\cite{gemma4}. The ranking of methods is identical under all three judges, and each method's score varies by at most $1.8$. Notably, the same-family judge scores Ours (Qwen3VL) at $53.1$, between the two independent judges ($54.3$ and $52.5$), showing no sign of same-family score inflation. Moreover, generation (greedy decoding with fixed frames and prompts) and judging (a single forward pass per triplet) are both deterministic, ensuring reproducibility.

\noindent\textbf{Error analysis.} Table~\ref{tab:error} categorizes false positives (triplets rejected by the judge) and false negatives (missed ground-truth relations). The false positives of OED spread across temporal, relation, and other (\textit{e.g.}, entity) errors, whereas $69.5$--$74.6\%$ of ours are correct triplets with imprecise temporal intervals. This suggests that MLLM-based DSGG is substantially more semantically consistent, and its main remaining weakness is temporal boundary estimation, which is bounded by the dense temporal localization ability of MLLMs. Meanwhile, our models disproportionately miss short and less informative relations: unlike traditional models that over-generate, MLLMs produce fewer spurious edges at the cost of coverage on less important triplets. Consistently, our SGQA gains are largest on event-level and multi-hop questions, which require multi-edge coherence (Table~\ref{tab:mcq}).

\noindent\textbf{Comparison with zero-shot frontier MLLMs.} Open-source MLLMs perform poorly on zero-shot DSGG (Figure~\ref{fig:zs_recall_comparison}). To test whether proprietary MLLMs remove the need for finetuning, we evaluate GPT-5.6 Luna~\cite{gpt5_6luna} and Gemini 3.5 Flash-Lite zero-shot (Table~\ref{tab:zs_frontier}). Both achieve high precision, as \emph{MLLM-as-a-Judge} credits valid triplets beyond the annotations, yet their margin over ours is modest; most of their errors are correct triplets with imprecise temporal intervals. Their recall and nDCG are substantially lower (\textit{e.g.}, $18.4$/$18.2$ \textit{vs.} $47.6$ nDCG@10), as their predictions are misaligned with the target relation distribution; frontier MLLMs thus do not remove the need for finetuning.

\noindent\textbf{Open-vocabulary DSGG.} This work follows the closed-vocabulary setting of mainstream DSGG benchmarks. As a preliminary step beyond it, we adopt the OpenVidVRD~\cite{openvidvrd} open-vocabulary protocol on VidVRD (Table~\ref{tab:openvoc}). Ours (Qwen3VL, 50\%) improves every metric over the strongest baseline by a wide margin, \textit{e.g.}, SGDet mAP from $17.95$ to $32.13$ and SGCls mAP from $18.25$ to $37.63$. Training on $50\%$ of the data outperforms using the full set: with a small, fixed vocabulary, longer training overfits the MLLM to it and hurts generalization. Faithfully evaluating open-vocabulary capacity thus calls for open-world, expressive DSGG datasets, which we leave for future work. Additional analysis and discussion are provided in Appendix~\ref{sec:discussion}.

\section{Conclusion}
\label{conclusion}

In this work, we re-examine Dynamic Scene Graph Generation (DSGG) from the perspective of practical utility in the era of Multimodal Large Language Models (MLLMs). We identify key limitations of existing DSGG approaches, and introduce a set of practicability-oriented evaluation metrics that better reflect the usefulness of generated scene graphs in downstream tasks. Building on these insights, we explore the use of MLLMs for DSGG and propose several key designs. Extensive experiments across multiple datasets demonstrate that our approach not only achieves SOTA recall performance but also produces more informative, balanced, and practically useful scene graphs.

\nocite{kang2025lp}
\nocite{kang2026causality}
\nocite{deng2025covis}
\bibliographystyle{plain}
\bibliography{main}

\appendix
\clearpage

\section{SGQA Dataset Construction} 
To evaluate the downstream utility and semantic fidelity of generated video scene graphs (SGs), we introduce \textbf{SGQA}, a diagnostic benchmark comprising multiple-choice questions (MCQs) grounded in scene graph structures. We utilize Gemini 2.5 Pro~\cite{gemini2_5} to curate these questions, conditioned on both the video and the ground-truth dynamic scene graphs represented in our proposed Temporal Relation Set (TRS) format: $\langle \textit{sub}, \textit{rel}, \textit{obj}, t_{\text{start}}, t_{\text{end}} \rangle$. Specifically, we define five core reasoning dimensions to assess the structural and temporal completeness of the SGs:

\begin{itemize}
    \item \textbf{Object/Action Identification:} Measures the ability to verify the presence of fine-grained entities or subtle interactions that exceed simple label matching. \textit{E.g., ``Which specific object is being manipulated during the primary interaction?''}
    \item \textbf{Spatial Reasoning:} Evaluates the understanding of 3D relative positioning and orientation between entities within the scene. \textit{E.g., ``What is the relative position of the person with respect to the shelf at the start of the clip?''}
    \item \textbf{Temporal Reasoning:} Tests the capacity to track state transitions, sequential dependencies, and the chronological order of relations. \textit{E.g., ``What action did the person perform immediately before interacting with the cupboard?''}
    \item \textbf{Event-Level Reasoning:} Requires high-level abstraction of discrete triplets into a coherent understanding of intent or activity context. \textit{E.g., ``Which overarching activity best summarizes the sequence of interactions observed?''}
    \item \textbf{Multi-hop Reasoning:} Serves as a complexity stress test requiring the model to traverse multiple relational dependencies and nodes simultaneously to reach a conclusion.
\end{itemize}

\begin{figure}
    \centering
    \includegraphics[width=\linewidth]{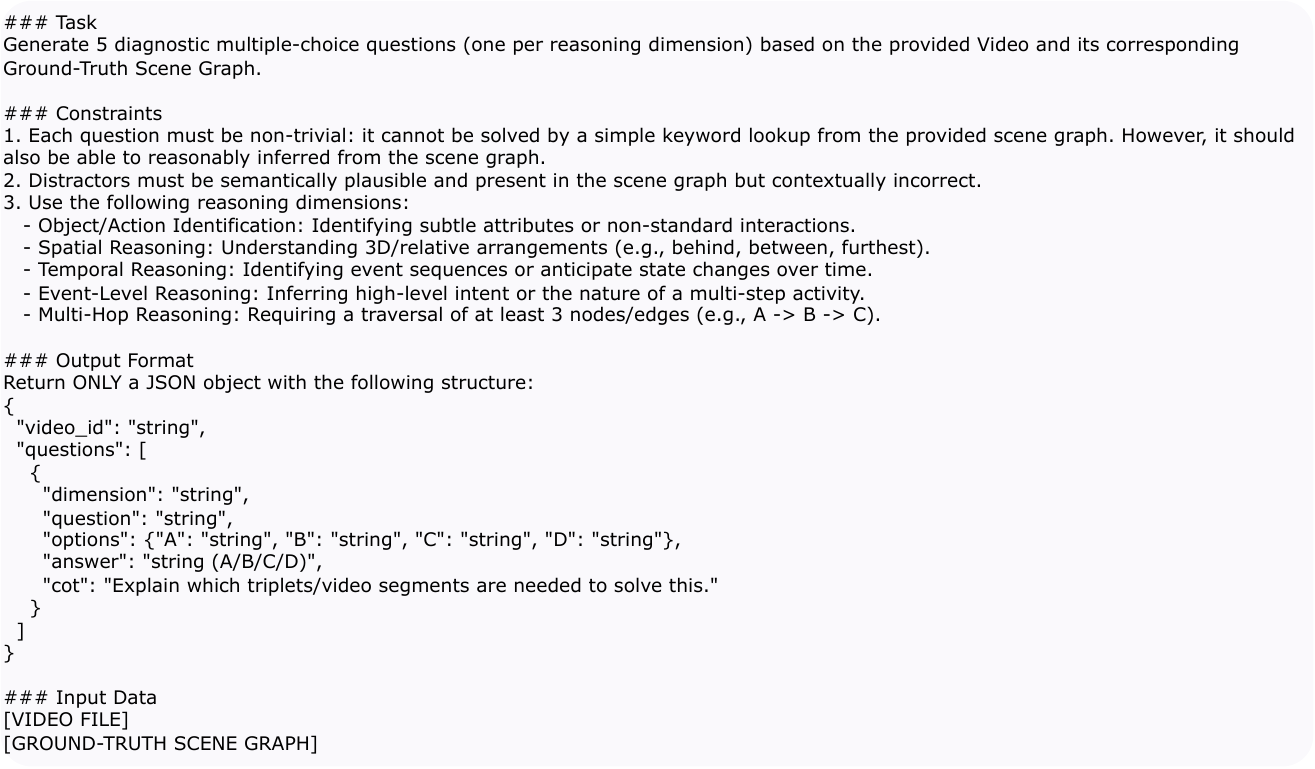}
    \caption{Prompt for generating the proposed SGQA dataset.}
    \label{fig:dataset_gen_prompt}
\end{figure}

For each dimension, we randomly sample 200 video clips from Action Genome~\cite{ji2020actiongenome}, with each clip represented by 8 keyframes, totaling 1,000 diagnostic samples. To ensure the benchmark is sufficiently challenging, we prompt the generator to craft "hard" questions with semantically plausible distractors that cannot be resolved by simply traversing the scene graph. The prompt used for generating dataset is shown in Figure~\ref{fig:dataset_gen_prompt}.

\textbf{Evaluation.} To evaluate the performance of DSGG with our constructed MCQ, we first take the top-$20$ generated scene graph triplets and convert into TRS format. Then we use Qwen3-8B~\cite{qwen3} as the fixed evaluator to answer MCQs based on generated SGs.

\section{Implementation Details}
\label{sec:impl_details}

\textbf{MLLM Backbone.} We adopt two state-of-the-art MLLMs: InternVL3.5~\cite{internvl3.5} and Qwen3VL~\cite{qwen3vl} with 8B model size as the scene graph generator. We use Qwen3VL-2B for calculating informativeness score. We use Jina V3 text embedding model~\cite{sturua2024jinaembeddingsv3multilingualembeddingstask} for calculating diversity score. For \emph{MLLM-as-a-Judge}, we choose the \texttt{Qwen/Qwen3-VL-30B-A3B-Instruct-FP8} model, balancing performance and efficiency. For the cross-judge analysis (Table~\ref{tab:judge}), we additionally use Gemini 3.5 Flash-Lite~\cite{gemini3_5flashlite} and Gemma 4 26B-A4B~\cite{gemma4} as judges.

\textbf{Frame Sampling.} For frame sapling, we use consecutive frames whenever there is scene graph annotation, and each sample consists of 8 frames.

\textbf{SGRet setup.} We adopt Qwen3VL-Embedding-2B~\cite{qwen3vl_embedding} as the multimodal embedding model for embedding both textual scene graph and the video clip. We randomly sample 1000 video clips as the retrieval pool. 

\textbf{Grounding with Off-the-Shelf Detector.} To perform accurate grounding with an off-the-shelf open-vocabulary detector (\textit{e.g.} SAM 3~\citep{sam3}), we use a combination of textual labels and points generated by the MLLM scene graph generator. Specifically, we first pass each triplet as referring expression (\textit{e.g.} \texttt{person holding cup}) to the detector. In case when there are multiple predictions, we select the mask whose centroid is closest to the generated point.

\begin{figure}
    \centering
    \includegraphics[width=\linewidth]{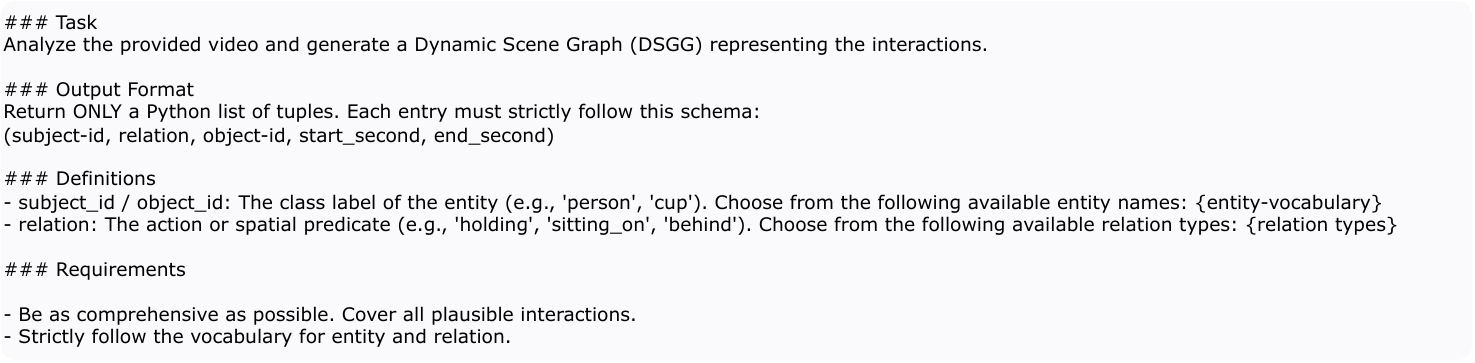}
    \caption{Prompt for zero-shot DSGG experiments.}
    \label{fig:zs_prompt}
\end{figure}

\textbf{Prompts.} The prompt for zero-shot DSGG experiment is shown in Figure~\ref{fig:zs_prompt}. For \emph{MLLM-as-a-Judge}, we use the prompt: 

``\texttt{Given the video and the following subject-relation-object triplet, determine if the triplet is valid based on the video. Answer Yes or No.}''. 

For the MLLM scene graph generator, we adopt the prompt: 

``\texttt{Given the video, generate a python list of tuples in the form of ((subject-id, start-x, start-y), relation, (object-id, start-x, start-y), start\_time, end\_time).}''

\section{Training details}
\label{sec:trainingdetails}
We apply LoRA~\cite{lora} finetune for all experimented backbone MLLMs with two Nvidia H100 GPUs. The LoRA rank is set to 32 with an alpha value of 64 across all experiments. We train all models for a single epoch. The learning rates are set to be $4 \times 10^{-5}$ for Qwen3-VL and InternVL3.5. Batch sizes are set to 1 with gradient accumulation steps equal to 8.

% \section{Importance-Aware Finetuning with CLIP score}

% \paragraph{Details for Triplet Importance Calculation}
% We calculate Triplet Importance ($TI$) based on a combination of triplet \textit{informative-ness} $T_I$ and triplet \textit{diversity} $T_D$. Given an image $I$ and $i$-th corresponding triplet $T_i$, $TI = \lambda T_I + (1-\lambda) T_D$ with:

% \begin{align*}
%     T_I &= \text{cos}(Enc_i(I), Enc_t(T_j)) \\
%     T_D &= \begin{cases}
%         1, & \text{if } i = 1 \\
%         \displaystyle 1 - \frac{1}{i-1}\sum_{k=1}^{i-1}{\text{cos}(Enc_t(T_k), Enc_t(T_i))}, & \text{otherwise}
%     \end{cases} \\
% \end{align*}

% where $\text{cos}(\cdot, \ \cdot)$ denotes the cosine similarity function, $Enc_i$ and $Enc_t$ denotes CLIP's image and text encoder. We experimentally set $\lambda = 0.75$ to give a higher weight for $T_I$, as $T_D$'s scale of change is often larger than $T_I$'s, which outweighs the former.

\section{Additional results}

\begin{wrapfigure}{r}{0.54\linewidth}
    \centering
    % \vspace{-10pt}
    \includegraphics[width=\linewidth]{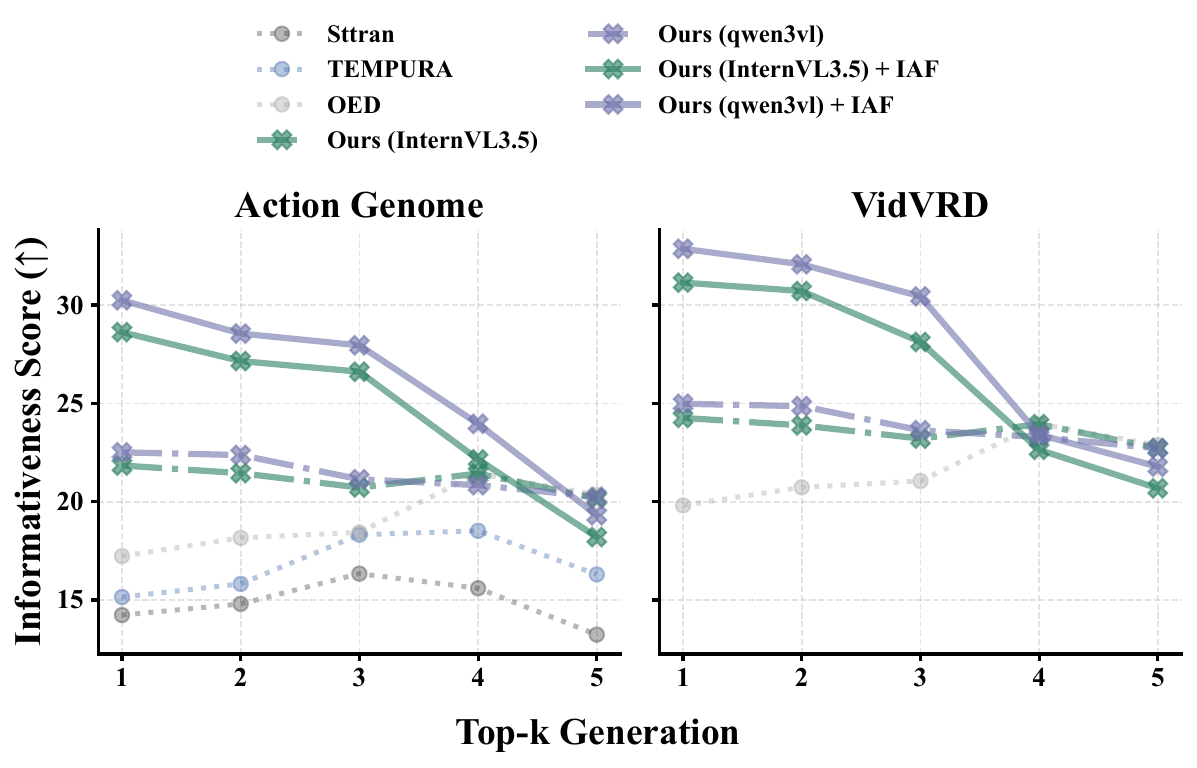}
    \caption{Top-5 triplets' informativeness score generated by STTran~\cite{sttran}, TEMPURA~\cite{tempura}, ours, and ours with Importance-Aware Finetuning (IAF)~\cite{ji2020actiongenome} on Action Genome and VidVRD dataset.}
    \vspace{-10pt}
    \label{fig:kth_score}
\end{wrapfigure}

\paragraph{Top-k generation's informativeness score.} 
Figure~\ref{fig:kth_score} compares the informativeness scores of the top-$k$ generated triplets across different methods on Action Genome and VidVRD. Overall, MLLM-based approaches consistently produce more informative relations than existing DSGG models. Traditional methods such as STTran and TEMPURA tend to generate triplets with significantly lower informativeness scores, suggesting that their high-confidence predictions often correspond to trivial or redundant relations. In contrast, our MLLM-based models generate relations that better capture the key semantics of the video. Furthermore, applying Importance-Aware Finetuning (IAF) further improves the ranking quality of generated triplets, especially for the top-ranked predictions. This effect is most noticeable at small $k$, where IAF encourages the model to prioritize highly informative relations earlier in the generation process. As $k$ increases, the informativeness scores gradually decrease for all methods, reflecting the diminishing semantic value of lower-ranked triplets. Nevertheless, our approach consistently maintains higher informativeness scores across both datasets, demonstrating the effectiveness of the proposed ranking-aware training strategy.

\paragraph{SGQA Performance Across Question Types.}Table~\ref{tab:mcq} reports the SGQA accuracy across different question categories. Our MLLM-based approaches significantly outperform prior DSGG methods across all question types. In particular, \textbf{Ours (Qwen3-VL)} achieves the best overall accuracy of $48.9\%$, improving over the strongest baseline (OED) by $+12.8$. The improvements are consistent across all reasoning categories, including object identification, spatial reasoning, temporal understanding, event comprehension, and multi-hop reasoning. Notably, the largest gains are observed on the more complex reasoning tasks such as \emph{Multi-hop} and \emph{Event} questions, where our method improves by up to $+19.4$ and $+13.1$, respectively. These results indicate that the scene graphs generated by our MLLM-based framework capture richer semantic and temporal information, enabling more effective reasoning over video content.

\begin{table}[t]
\centering
\caption{Performance across different MCQ question types. Task abbreviation is as follow: \textbf{Spatial}: spatial understanding; \textbf{Temporal}: temporal understanding; \textbf{Event}: event understanding; \textbf{Multi-hop}: multi-hop reasoning.}
\resizebox{0.85\linewidth}{!}{
\begin{tabular}{lcccccc}
\toprule
\textbf{Model} & \textbf{Obj} & \textbf{Spatial} & \textbf{Temporal} & \textbf{Event} & \textbf{Multi-hop} & \textbf{Overall} \\
\midrule
STTran       & 26.5 & 25.8 & 25.1 & 25.5 & 24.8 & 25.5 \\
TEMPURA      & 36.9 & 33.6 & 27.9 & 33.6 & 25.7 & 31.5 \\
OED          & 44.1 & 40.2 & 28.5 & 40.3 & 27.2 & 36.1 \\
\midrule
\rowcolor{lblue}
\textbf{Ours} (InternVL3.5)  & 52.4\,\inc{8.3} & 51.3\,\inc{11.1} & 38.4\,\inc{9.9} & 52.6\,\inc{12.3} & 44.4\,\inc{17.2} & 47.8\,\inc{11.7} \\
\rowcolor{lblue}
\textbf{Ours} (Qwen3-VL)     & \textbf{53.2}\,\inc{9.1} & \textbf{51.8}\,\inc{11.6} & \textbf{39.6}\,\inc{11.1} & \textbf{53.4}\,\inc{13.1} & \textbf{46.6}\,\inc{19.4} & \textbf{48.9}\,\inc{12.8} \\
\bottomrule
\end{tabular}
}
\label{tab:mcq}
\end{table}

\begin{figure*}[t]
    \centering
    \includegraphics[width=\textwidth]{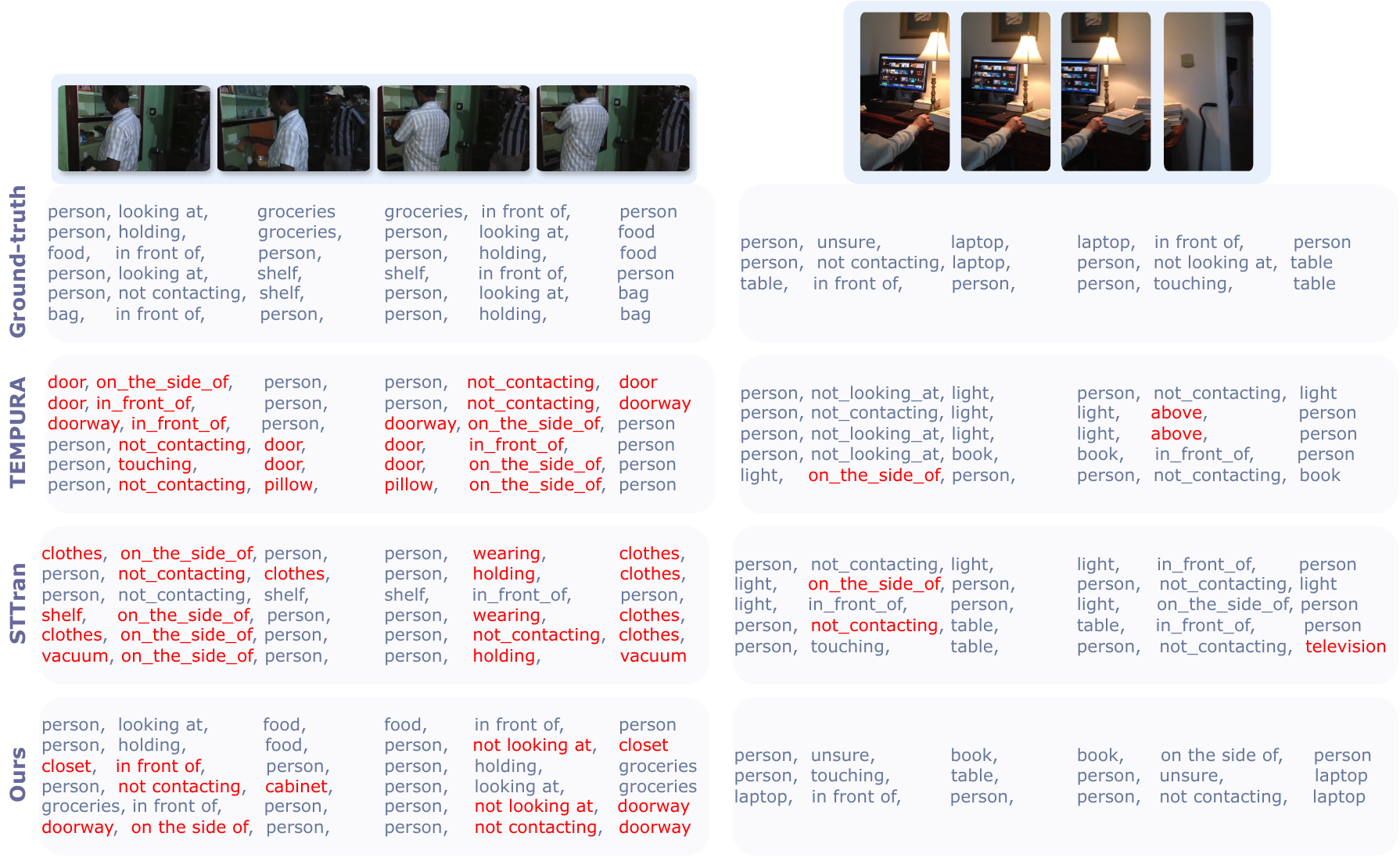}
    \caption{Qualitative comparison between Tempura, STTran, and Ours. Predictions/objects highlighted in \textcolor{red}{red} represent false predictions. Triplets are displayed (top-down) in the order of their model predictions (sorted by probability for STTran and TEMPURA, or by their order of generation.}
    \label{fig:sttranvstempuravsllmdsgg_comparison}
    % \vspace{-0.15in}
\end{figure*}

\begin{figure*}[t]
    \centering
    \includegraphics[width=\textwidth]{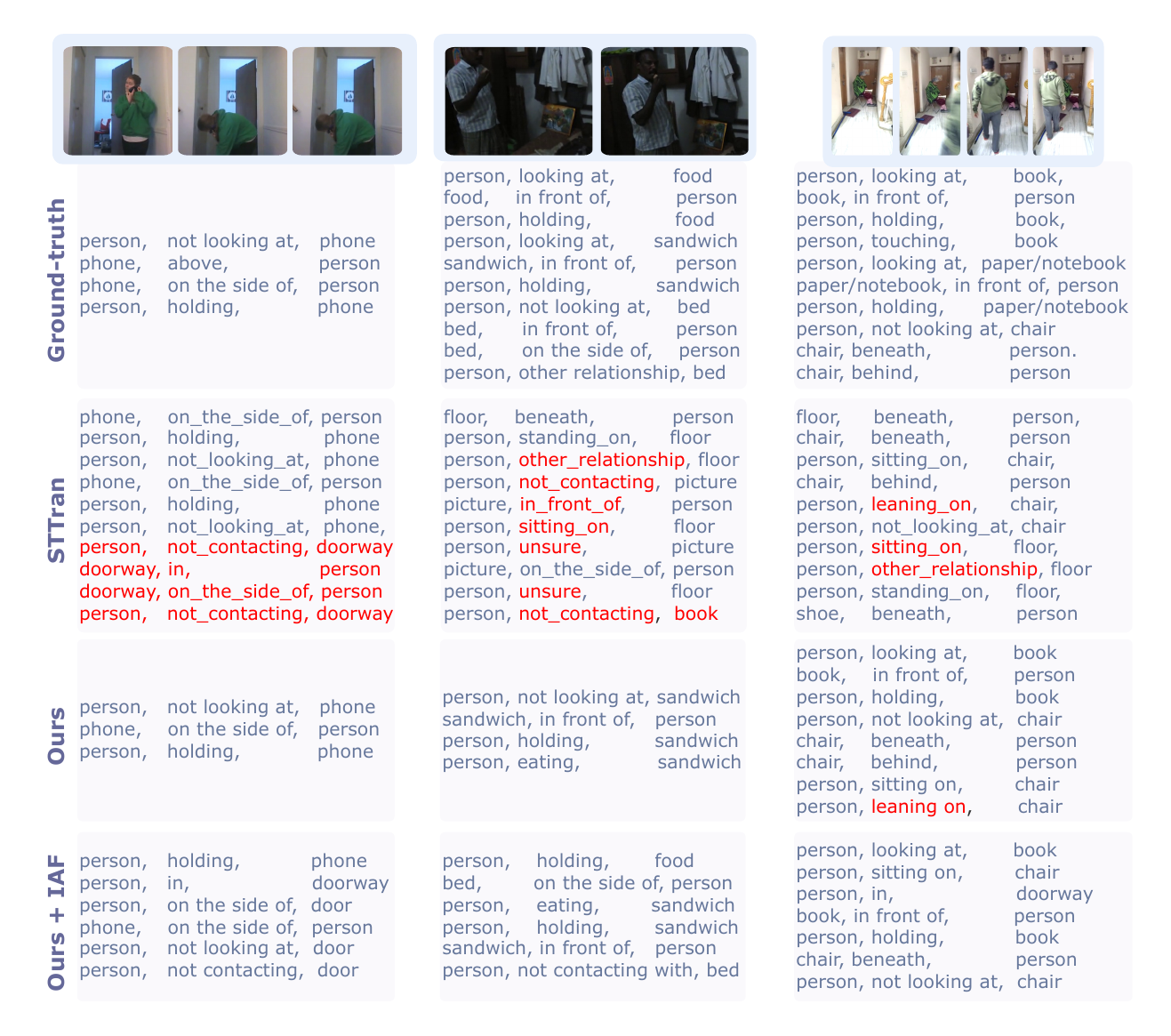}
    \caption{Qualitative examples of generations from traditional DSGG method (STTran~\cite{sttran}), versus ours, and ours with Importance-Aware Finetuning (IAF). Predictions/objects highlighted in \textcolor{red}{red} represent false predictions. Triplets are displayed (top-down) in the order of their model predictions (sorted by probability for STTran and TEMPURA, or by their order of generation.}
    \label{fig:TopkQualitativeResultsAG}
    % \vspace{-0.15in}
\end{figure*}

\paragraph{Qualitative comparison between LMM and Previous methods.}Figure~\ref{fig:sttranvstempuravsllmdsgg_comparison} and Figure~\ref{fig:TopkQualitativeResultsAG} compare scene graph generated from STTran~\cite{sttran}, TEMPURA~\cite{tempura}, Ours, and Ours with Importance-Aware Finetuning (IAF). We can observe STTran and TEMPURA are able to generate more triplets but also exhibit a higher rate of false positives, including the generation of context-based triplets—instances where triplets are produced even when the corresponding objects are absent from the scene. In contrast, Ours-based results are noticeably more accurate. We can observe that as the number of predictions increases, STTran and TEMPURA tend to generate more false positives, leading to a decline in precision at higher values of $K$. In contrast, Ours effectively aligns triplet generation with the scene context, maintaining higher precision. Meanwhile, we can also observe from both Figure~\ref{fig:sttranvstempuravsllmdsgg_comparison} and Figure~\ref{fig:TopkQualitativeResultsAG} that previous methods like STTran and TEMPURA tends predict prevalent objects and relations such as "floor" and "not looking at" as the highest confidence entities and relations. However, in the given examples, these entities and relations are either not visible in the scene, or are not important to the primary activity. On the other hand, Ours with IAF is able to generate the most relevant triplets first. For instance, for the right most sample of Figure~\ref{fig:TopkQualitativeResultsAG}, where the person is sitting on a chair while reading a book, while STTran predicts $ \langle \textit{floor}, \textit{beneath}, \textit{person} \rangle $, Ours is able to predict $\langle \textit{person}, \textit{looking at}, \textit{cabinet} \rangle$, which is more relevant to the person's action. Meanwhile, by training with IAF, the results also involve less redundant triplets at a rank position. Take the same example for instance, while STTran predicts redundant triplets: $ \langle \textit{chair}, \textit{beneath}, \textit{person} \rangle $, $ \langle \textit{person}, \textit{sitting on}, \textit{chair} \rangle $, $ \langle \textit{chair}, \textit{behind}, \textit{person} \rangle $, Ours $+$ IAF is able to predict distinct distinct triplets up to a rank position.

\section{Additional Analysis and Discussion}
\label{sec:discussion}

\paragraph{Evaluation protocol and reproducibility.} Both scene graph generation and \emph{MLLM-as-a-Judge} are deterministic in our evaluation. For generation, we use greedy decoding with fixed frames and prompts. For judging, we run a single forward pass per triplet with the fixed prompt in Appendix~\ref{sec:impl_details} and extract the logit of the \texttt{"Yes"} token. Replacing the default Qwen3-VL-30B-A3B judge with Gemini 3.5 Flash-Lite or Gemma 4 26B-A4B preserves the ranking of all evaluated methods, with a maximum within-method difference of $1.8$ in $P@10$ (Table~\ref{tab:judge}). We will release the code, datasets, and evaluation pipeline so that future methods can be evaluated under the same protocol.

\paragraph{Restricting predictions to the target taxonomy.} Although MLLMs and SAM~3 are open-vocabulary models, we find that finetuned MLLMs are faithful to the target vocabulary. Soft-matching the generated labels to the dataset vocabulary with a text-embedding model brings only a marginal ($<1\%$) gain in recall, as the finetuned MLLMs rarely generate novel labels that are synonyms of existing categories. When they do generate truly novel labels, these predictions are assessed by the \emph{MLLM-as-a-Judge} precision. SAM~3 is applied only after entity and relation generation and does not alter the predicted labels: for each predicted triplet, we pass the triplet as a referring expression (\textit{e.g.}, \texttt{person holding cup}) to SAM~3 and, when multiple masks are returned, select the mask whose centroid is closest to the subject/object point predicted by the MLLM (Appendix~\ref{sec:impl_details}).

\paragraph{Identity association and occlusion.} The MLLM assigns an identifier to each entity (\textit{e.g.}, \texttt{person-0} in Figure~\ref{fig:diagram}), which associates the same entity across different triplets. We do not explicitly model occlusions or entities that temporarily leave the frame. Instead, such entities are inferred implicitly through the holistic video understanding of the MLLM, while SAM~3 tracks them through partial or short-term occlusions with its memory. This is another advantage of the reason-then-locate paradigm: locate-then-reason methods operate on detected regional features and are therefore inherently limited when entities are invisible; accordingly, the standard DSGG evaluation protocol excludes occluded objects (\textit{e.g.}, by filtering invisible objects out of the ground truth).

\paragraph{Comparison with traditional DSGG models.} Our MLLM-based models benefit from the large-scale pretraining of MLLMs and SAM~3, whereas most traditional DSGG models are trained from scratch, which leads to an asymmetry in model scale and pretraining data. However, leveraging pretrained foundation models is an established practice in scene graph generation: $\text{VS}^3$~\cite{vs3} exploits pretrained vision-language models for image SGG, and LASER~\cite{laser} and ESCA~\cite{esca} distill the knowledge of commercial MLLMs to train DSGG models. Moreover, MLLMs do not naturally excel at DSGG: open-source MLLMs perform poorly on zero-shot DSGG (Figure~\ref{fig:zs_recall_comparison}), and even frontier proprietary MLLMs used zero-shot attain substantially lower recall and nDCG than finetuned models (Table~\ref{tab:zs_frontier}). Rather than proposing another specialized module within the traditional DSGG framework, our goal is to analyze the failure modes of existing DSGG models, to establish practical evaluation metrics for DSGG, and to provide a systematic study of MLLM-based DSGG that can serve as a solid foundation for future work.

\paragraph{Open-vocabulary evaluation protocol.} For the results in Table~\ref{tab:openvoc}, we follow the open-vocabulary protocol of OpenVidVRD~\cite{openvidvrd} on ImageNet-VidVRD~\cite{vidvrd}, which splits object and predicate categories by frequency into base (common) and novel (rare) categories as in RePro~\cite{repro}. Models are trained on base categories, and results are reported on the test set under the Novel split, which includes all object categories and the novel predicate categories. We report mAP and $R@50/100$ for SGDet and SGCls, where SGCls uses ground-truth trajectories, and a predicted triplet is considered correct if it matches a ground-truth triplet with a trajectory IoU above $0.5$. The $50\%$ and $100\%$ settings denote the fraction of training data used for finetuning, and baseline results are taken from~\cite{openvidvrd}.

\paragraph{Details of the error analysis.} For false positives (Table~\ref{tab:error}), we categorize the predicted triplets rejected by \emph{MLLM-as-a-Judge} into correct triplets with incorrect temporal intervals (Temp.), correct subjects and objects with incorrect relations (Rel.), and other errors such as incorrect entities (Other). For false negatives, we report the fraction of missed ground-truth relations whose informativeness score (Section~\ref{sec:methods}) or duration is below the mean; the two attributes are independent and can overlap. Beyond these statistics, a qualitative audit of SGRet failures shows that traditional models repeatedly emit the same frequent, low-information relations across distinct videos, making their graph embeddings less discriminative, whereas the residual failures of our models more often involve attributes and fine-grained relations outside the dataset taxonomy. Qualitative comparisons of false predictions are shown in Figures~\ref{fig:sttranvstempuravsllmdsgg_comparison} and~\ref{fig:TopkQualitativeResultsAG}.

\section{Limitation}

While this work studies adopting MLLMs for direct DSGG, the current scope mainly focuses on the closed-vocabulary setting. Although our preliminary open-vocabulary results are promising (Table~\ref{tab:openvoc}), faithfully evaluating open-vocabulary, in-the-wild DSGG requires open-world and expressive DSGG datasets, which we leave for future work. In addition, temporal boundary estimation remains the main error source of MLLM-based DSGG (Table~\ref{tab:error}) and is hard to address with lightweight finetuning on small-scale DSGG datasets, which motivates improving the dense temporal localization ability of MLLMs. Finally, since the current pipeline mainly considers short video clips (e.g. 8 frames), it will be helpful to extend MLLM-based DSGG to long-video understanding, considering downstream applications of DSGG (e.g. long-video summarization and reasoning).

% \newpage
% \clearpage
% \input{checklist.tex}
\end{document}